\documentclass[12pt,a4]{article}
\usepackage{geometry}
\usepackage{multirow}
\usepackage{graphicx}
\usepackage{graphics} 
\usepackage{epsfig}
\usepackage{amsmath, amsthm, amssymb, amsfonts, enumerate}
\usepackage{blindtext}
\usepackage{float}
\usepackage{colortbl}
\usepackage[english]{babel}
\usepackage{authblk}
\usepackage[colorlinks=true, pdfstartview=FitV,breaklinks=true, 
linktocpage = true, citecolor = magenta]{hyperref}
\usepackage{algorithm}

\usepackage[noend]{algpseudocode}
\usepackage{times} 
\usepackage{url}
\usepackage{booktabs}
\usepackage{array}

\usepackage{sidecap}
\sidecaptionvpos{figure}{t}

\usepackage{sidecap}
\sidecaptionvpos{figure}{t}

\usepackage[toc,page]{appendix}

\usepackage{subcaption}

\newtheorem{lemma}{Lemma}
\newtheorem{proposition}{Proposition}
\newtheorem{theorem}{Theorem}

\theoremstyle{definition}

\theoremstyle{definition}
\newtheorem{remark}{Remark}
\theoremstyle{definition}

\algrenewcommand\algorithmicrequire{\textbf{Input:}}
\algrenewcommand\algorithmicensure{\textbf{Output:}}

\usepackage{lineno}

\begin{document}
	\title{Multiple Shooting Approach for Finding Approximately Shortest Paths for Autonomous Robots in  Unknown Environments in 2D}

	\author[1,2]{Phan Thanh An \thanks{E-mail: \texttt{thanhan@hcmut.edu.vn}}}
	\author[3,4]{Nguyen Thi Le}
	\affil[1]{\small Institute of Mathematical and Computational Sciences, Ho Chi Minh City University of Technology (HCMUT), 268 Ly Thuong Kiet Street, District 10, Ho Chi Minh City, Vietnam}
	\affil[2]{\small  Vietnam National University Ho Chi Minh City, Linh Trung Ward, Thu Duc District, Ho Chi Minh City, Vietnam}
	\affil[3]{\small People's Security Academy, 125 Tran Phu Street, Ha Dong District, Hanoi, Vietnam}
	\affil[4]{\small Institute of Mathematics,  Vietnam Academy of Science and Technology (VAST), 18 Hoang Quoc Viet Road, 
		Hanoi, Vietnam}
	\date{}
	
	\maketitle
	\newenvironment{mylisting}
	{\begin{list}{}{\setlength{\leftmargin}{1em}}\item\normalsize\bfseries}
		{\end{list}}
	\newenvironment{mytinylisting}
	{\begin{list}{}{\setlength{\leftmargin}{1em}}\item\scriptsize\bfseries}
		{\end{list}}

	\maketitle
	
	\begin{abstract}
		An autonomous robot with a limited vision range finds a path to the goal  in an unknown environment in 2D avoiding polygonal obstacles.
		In the process of discovering the environmental map,
		the robot has to return to some positions marked previously, the regions where the robot   traverses to reach that position are defined as sequences of bundles of line segments.
		This paper presents a novel  algorithm for finding approximately  shortest  paths along the sequences of bundles of line segments
		based on  the method of multiple shooting.
		Three factors of the
		approach including bundle partition, collinear condition, and update of shooting points are
		presented.
		We then show that
		if the collinear condition holds, the exact shortest path of the problem
		is  determined, otherwise, the sequence lengths of paths obtained by the
		update of the method converges.
		The algorithm is implemented in Python and some numerical examples show
		that the running time of path-planing for autonomous robots using our method is faster than that using the rubber band technique of Li and Klette in \textit{Euclidean Shortest Paths, Springer},  53--89 (2011).

	\end{abstract}
	
	\medskip
	{\bf Keywords:} Autonomous robot; sequence of line segments; multiple shooting; shortest path.

	\section{Introduction}\label{Introduction}%
	
	%
	%
	%

	Finding shortest paths is a natural geometric optimization problem  with many applications in different fields such as geographic information systems (GIS), robotics, computer vision, etc. (see ~\cite{Li2011,Mitchell2000}). 
	A case of the geometrical shortest path problem is to compute the shortest path joining two
	points along a set of line segments in 2D as follows:
	%
	
	\begin{itemize}
		\item[] \textit{Given two points and an ordered set of line segments in 2D, find the shortest path joining these
			points such that it passes through these line segments in the given order.}
	\end{itemize}
		We call the problem the SP problem for  line segments. If line segments are pairwise disjoint, it is called the SP problem for pairwise disjoint line segments. 
		Consider the SP problem for bundles of line segments as follows:
		\begin{itemize}
			\item[(P)] \textit{Given two points and a sequence of bundles of line segments in 2D in which these bundles do not intersect each other, find the shortest path joining two points along the sequence
				of bundles of line segments,
			}
		\end{itemize}	
		where the terms \textit{path along a sequence of bundles of line segments} and \textit{two bundles not intersecting each other} are described in detail in Section~\ref{sect:preliminaries}.
		Problem (P) is  also a case of the SP problem for line segments  and vice versa. However, the SP problem for bundles of line segments is more general than the SP problem for pairwise disjoint line segments. 
	
	By the convex optimization approach, Polishchuk and Mitchell~\cite{Mitchell2000} used a second-order cone program (SOCP) to solve the SP problem for arbitrary line segments, but this method is not popular since the authors stated that using SOCP is considered a “black box” without any knowledge of its internal workings.	
		There is currently no geometric algorithm available that can solve the SP problem for arbitrary line segments.  Nevertheless, exact and approximate solutions can be obtained under certain specific conditions of line segments~\cite{Lee1984, Li2011, Trang2017}. 
			Unfortunately, these algorithms are not applicable to the SP problem for bundles of line segments. 
		In particular, if the line segments of a sequence of bundles can be sorted as a list of diagonals of adjacent triangles, and then the funnel technique of Lee and Preparata~\cite{Lee1984} can be utilized to get the exact shortest path, whereas the domain containing the paths that we consider is not a triangulated polygon. 
		Therefore, applying the funnel technique is not impossible for the SP problem for bundles of line segments (see Appendix~\ref{apdx:example}).
		When dealing with the SP problem for pairwise disjoint line segments, one can use iterative algorithms using approximate approaches of Li and Klette~\cite{Li2011} and Trang et. al.~\cite{Trang2017}.
		However, these algorithms are not appropriate for non-pairwise disjoint line segments,
		such as when the set of line segments is a sequence of bundles (see Appendix~\ref{apdx:example}).
		%
		%
		%
		An et al.~\cite{BinhAnHoai2021} 
		used the rubber band technique of Li and Klette~\cite{Li2011} (with some modifications) 
		to solve approximately the SP problem for bundles of line segments.
		We know that the  rubber band technique of Li and Klette~\cite{Li2011} is applied for the SP problem for pairwise disjoint line segments, whereas two line segments of one bundle share a common point as the vertex of the bundle. 
	 In  the implementation, however, An et al.~\cite{BinhAnHoai2021} trimmed line segments of each bundle by shortening a sub-segment of the length of $\epsilon$ from the vertex to guarantee that all line segments within the bundles are pairwise disjoint,
		where $\epsilon$ is a very small positive number. 
		In theory, Li and Klette's technique still works. However, during execution, if $\epsilon$  is set to a small value,  there are some  specific cases where the algorithm was halted at certain iterations due to floating-point rounding errors. On the other hand, increasing $\epsilon$ to a larger value leads to an unsatisfactory solution. As a result, this approach fails to fully address the limitations of Li and Klette's technique.
		A question arises ``Is there an efficient algorithm for the SP problem for bundles of line segments without cutting off line segments?''
	%
	
	
	The geometric shortest path problem plays an important role in the path planning of autonomous robots  as one of  the major aspects of  robotics, see ~\cite{Brooks1985,Hsu2007,Lad2004,Warren1993}.
	%
	There are two approaches: global and local path
	planning according to the known level of environmental information.
	Global path planning is used in fully known environments, whereas local path planning requires local map information obtained when the robot moves in unknown environments with unknown or dynamic obstacles, see~\cite{Sedighi2004,Siegwart2011}.

Let us consider a problem of local path planning in which a robot  with a limited vision range moves in an environmental map in 2D avoiding obstacles that are polygons, see~\cite{BinhAnHoai2021,Kunvhev2006}. 
Notice that the robot does not have all information about obstacles except for the coordinates of starting and ending points (see Figure~\ref{fig:vision-range}).
Avoiding  obstacles requires that paths do not intersect with the interior of these obstacles. 
%
%
Apart from reaching the goal, saving time and reducing   lengths of motion paths are
also important.	
Let us assume the robot is at $q$ of the map. 
In the process of discovering the map, the robot might return to some position, says $p$, which does not belong to the current robot's vision. Although $p$ has not yet been traversed by the robot,  $p$ was marked and ranked in the past (for more detail, see Section~\ref{sect:autonomous_robot}).
%
Therefore, optimizing the path from $q$ to $p$  is established in this paper.
%
A	popular approach of path planning algorithms for robots is to use one available trajectory  for  returning. 
The trajectory is indeed the shortest path joining $p$ and $q$ in a graph $G$ whose nodes  are obtained from local information of the positions  that the robot has traversed in the past and whose edges are radial segments connecting nodes. 
The approach appears in path planning algorithms using  graph search-based  methods ~\cite{Brooks1985,Warren1993}, potential field algorithms~\cite{Glavaski2009,Rosell2005}, and roadmap algorithms~\cite{Hsu2007,Lad2004}.
To make the robot move effectively, An et al.~\cite{AnPatent2023,BinhAnHoai2021} proposed an autonomous robot operation algorithm in which they pay attention to finding a geometric shortest path to replace with the available trajectory obtained from  $G$. 
The  problem of optimizing paths for returning states that
\begin{itemize}
	\item[(P$^*$)] \textit{Find a  path (or shortest path)  joining $p$ and $q$ avoiding the obstacles, that is shorter than the 
		available trajectory obtained from the graph $G$.
	}
\end{itemize}
Section~\ref{sect:autonomous_robot} describes how the regions that the robot traverses from $p$ to $q$ are defined as sequences of bundles of line segments by that Problem (P$^*$) can be converted to a geometric problem.
Thus instead of solving
Problem (P$^*$), An et al. in~\cite{AnPatent2023,BinhAnHoai2021} dealt with the SP problem for bundles of line segments by the rubber band technique of Li and Klette with cutting off line segments, which are established in Section~\ref{sect:numerical_results}.
%

\begin{SCfigure}[][h]
	%
	\includegraphics[width=0.5\textwidth]{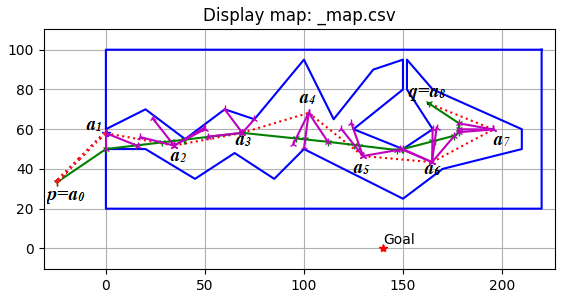}
	\caption{The available trajectory joining $p, a_1,a_2, \ldots, a_7$ and $q$ which is the shortest path  joining $p$ and $q$ in the graph $G$ (red dashed polyline) is replaced by  an approximate  shortest path (green solid polyline) joining $p$ and $q$ along the sequence of line segments obtained by using MMS.}
	\label{fig:path-obtained}
\end{SCfigure}

In this paper, we focus on solving approximately Problem (P)
and its application in  finding a path to the goal of the autonomous robot with a limited vision range avoiding obstacles.
The analytical and
geometrical properties of such paths including the existence, uniqueness, characteristics, and conditions for concatenation, are explored in~\cite{HaiAnHuyen2019}.
%
The method of multiple shooting (MMS), which was used successfully for determining shortest paths in polygon or polytope (see~\cite{AnHaiHoai2013,AnTrang2018,HoaiAnHai2017}), 
is proposed to find  shortest paths along  sequences of bundles of line segments. Note that MMS can be applied directly for (P), whereas Li and Klette's technique does not. An algorithm for path planning  for autonomous robots modified from that in~\cite{BinhAnHoai2021} is also addressed. 
If the robot moves from $q$ to $p$ in which $p$ is marked in the past and does not belong to the current robot's vision, MMS is called to find approximately  the   shortest path joining $p$ and $q$ along a sequence of bundles of line segments. It is a feasible path that is shorter than  the shortest path joining $p$ and $q$ in the graph $G$, see Figure~\ref{fig:path-obtained}.

%

We use MMS to compare with the method using Li and Klette's technique~\cite{Li2011} as a phase in path planning for autonomous robots	
introduced in~\cite{BinhAnHoai2021}. 	
We test on a set of different maps. The numerical experiments show that our
method averagely reduces the running time by about 2.88\% of the algorithm using Li and Klette's technique (Table~\ref{Table:time-results}).
%
Especially, when focusing
geometric shortest path problem, our algorithm outperforms the other significantly.
If   only considering the running time of solving sub-problems of finding approximately shortest paths along  sequences of line segments (not for the whole problem), the algorithm using MMS reduces by about 64,46\% of that using Li and Klette's technique. 
Besides, the total distance traveled by the robot  reduces significantly by about 16.57\%, as compared with the approach without replacing the available trajectories obtained from the graph $G$ by shorter paths for returning (Table~\ref{Table:length-results}).

%
The rest of the paper is organized as follows. Section~\ref{sect:preliminaries} presents preliminary
notions. 
In 
Section~\ref{sect:MMS},
an iterative algorithm based on MMS with three main  factors  is proposed (Algorithm~\ref{alg:main}) to find the shortest paths along sequences of line segments having common points.
Geometrically,  Algorithm~\ref{alg:main} gives multiple connections between two consecutive points of a set of points to obtain a path after each iteration.
Section~\ref{sect:correctness}  shows that if the stop condition of MMS holds, then the shortest  path is obtained (Proposition~\ref{prop:collinear-condition}).  Otherwise, the sequence of paths obtained by Algorithm~\ref{alg:main} is convergent by the length (Proposition~\ref{prop:update_path}) and Hausdorff distance (Theorem~\ref{theo:global solution}). 
%
Section~\ref{sect:refer_problem}  introduces an application of MMS for solving a phase of the problem of autonomous robots in unknown environments, which is referred to as the  geometric shortest path problem.
The proposed algorithm is implemented in Python and numerical results demonstrate the comparison with an algorithm using Li and Klette's technique in Section~\ref{sect:numerical_results}.
Some proofs for the correctness of the algorithm and examples showing the drawbacks of the previous algorithms
are arranged in the Appendices.

\section{Preliminaries}\label{sect:preliminaries}

We now recall some basic concepts and properties. For any points $x, y$ in the plane, we denote $[x,y] := \{ (1-\lambda)x + \lambda y: 0 \leq \lambda \leq 1 \}, \,
]x,y] := [x,y] \setminus \{ x \},$ and $]x,y[ := [x,y] \setminus \{ x, y \}.$
Let us denote $B(c,r) = \{ p \in \mathbb{R}^2 | \, \|p-c \| <r \}$, $\overline{B}(c,r) = \{ p \in \mathbb{R}^2 | \, \|p-c \| \le r \}$, and $\text{bd} \overline{B}(c,r) = \{ p \in \mathbb{R}^2 | \, \|p-c \| = r \}$, where $r>0$.
The fundamental concepts such as polylines, polygons, sequences of line segments, shortest paths, and their properties which will be used in this paper, can be seen in~\cite{An2017} and~\cite{HaiAnHuyen2019}.

In $\mathbb{R}^2$, a \textit{path} joining $p$ and $q$  is a continuous mapping $\gamma$ from an interval $[t_0, t_1] \subset \mathbb{R}$ to $\mathbb{R}^2$ such that $\gamma (t_0)=p$ and $\gamma (t_1)=q$.
A path $\gamma$ joining $p$ and $q$  is called a \textit{path along a sequence of  line segments} $e_1, e_2, \ldots, e_n$ if there is a sequence of
numbers $t_0 \le \bar{t_1} \le \bar{t_2} \le \ldots \le \bar{t_n} \le t_1$ such that $\gamma(\bar{t_i}) \in e_i$ for $i = 1, 2, \ldots, n$. 
%
We also call the image $\gamma([t_0, t_1])$ to be the path $\gamma$. All paths considered in the paper are polylines.
The \textit{shortest path} joining $p$ and $q$ along a sequence of  line segments $e_1, e_2, \ldots, e_m$, denoted by ${\rm SP}(p,q)$, is defined to be
the path of minimum length.

Let $\{ [a, b_1], [a, b_2], \ldots, [a, b_m] \}$ $(m \ge 1)$ be a sequence of   $m$  
distinct line segments in $\mathbb{R}^2$.
%
It is called a	\textit{bundle of $m$ line segments} with the \textit{vertex} $a$.
We  write the bundle whose vertex is $a$ simply the bundle $a$ when no confusion can arise.
%
A singleton point  is considered a degenerate bundle.
%
%
 We say that two bundles $a$ and $a'$ ($a \neq a'$)  \textit{intersect each other} if there are two line segments $[a, b]$ of $a$ and $[a', b']$ of $a'$  such that $[a,b] \, \cap \, [a',b'] \, \neq \emptyset$. Otherwise, we say that $a$ and $a'$ \textit{do not intersect each other}. 


Let us consider an ordered  set  of     bundles of line segments $a_1, a_2, \ldots, a_N$. It is called a \textit{sequence  of bundles of line segments}. 
We use the term a \textit{path along a sequence of bundles of  line segments}, which is  a path along a sequence of  line segments with the order of the bundles and that of the line segments in each bundle. 
In this paper, the term of a sequence of line segments/bundles of line segments is understood as a finite number of line segments/bundles of line segments in a given order. 
%


\section{Finding the shortest path joining two points along a sequence  of  bundles of line segments}
\label{sect:MMS}



%
This section introduces an iterative algorithm based on MMS to find approximately the shortest path joining two points along a sequence of bundles of line segments.
%
Let $\mathcal{F}$ be a sequence of bundles of line segments $a_0, a_1, \ldots, a_{N+1}$ ($N \ge 1)$, where $a_0=p,a_{N+1} =q$. 
Throughout the paper, assume that $\tau$ is a polyline joining $a_0, a_1, \ldots, a_{N+1}$.
We adopt the convention that a non-degenerate bundle $a_i$ consists of one or more line segments whose endpoints  are on $\tau$ or on the same side of $\tau$. A degenerate bundle is considered as consisting of one line segment whose endpoints are coincident. In this section, we only consider sequences of bundles of line segments in which these bundles do not intersect each other, and the path $\tau$ is a simple polyline.
For each non-degenerate bundle $a_i (1 \le i \le N)$, its line segments are sorted in order of the angles to be measured in the  order from the vector $\overrightarrow{a_i a_{i-1}}$ to the vector formed by $a_i$ and the remaining endpoints. The order of line segments in a non-degenerate bundle $a_i$ depends  on  $\overrightarrow{a_i a_{i-1}}$ rotating clockwise or counterclockwise to $\overrightarrow{a_i a_{i+1}}$ such that it scans through the region containing the bundle. 
Problem (P) is a case of the problem of finding the shortest paths along  sequences of line segments with the assumption that line segments might have common points. 
To deal with  (P), we use MMS (\cite{AnHaiHoai2013},~\cite{AnTrang2018},~\cite{HoaiAnHai2017}) with three factors as
follows:



\begin{itemize}
\item[(f1)] Partition  the sequence $\mathcal{F}$ of bundles of line segments into sub-sequences of bundles. 	Take an ordered set of cutting segments which are the last line segments of each sub-sequence.
We take  an initial shooting point in each cutting segment.
\item[(f2)] Construct a path along $\mathcal{F}$ joining  $p$ and $q$, formed by the set of shooting points. The path is the concatenation of the shortest paths joining two consecutive shooting points  along the corresponding sub-sequence.
A stop condition   is established at shooting points.
\item[(f3)] The algorithm enforces (f2) at all shooting points 
to check the stop condition. 
If the collinear condition is not met, an update of shooting
points will improves paths joining $p$ and $q$ along $\mathcal{F}$, which is formed by the set of new shooting points. 
\end{itemize}

The next parts will provide a detailed exposition of  these factors and the notions of cutting segments, and shooting points.

%
%

\subsection{Factor (f1): partition} 
\label{sect:partition}
Let $K$ be a natural number such that $1 \le K \le N$.
Partitioning of the sequence $\mathcal{F}$ of  bundles of line segments into  sub-sequences $\mathcal{F}_i$ of bundles of line segments and taking   so-called \textit{cutting segments} $\xi_i$, for $i=0,1, \ldots, K+1$ as by:

\begin{equation}
\label{eq:decomposition} 	\left\{ \begin{array}{lll}
	\mbox{+  }\mathcal{F}_i \text{ consists of a bundle or some adjacent bundles of }\mathcal{F}  \mbox{ for } i=0,1, \ldots, K. \\
	\mbox{+ }\xi_{i} \text{ is the last line segment of }\mathcal{F}_{i-1}, \mbox{ for } i=1,2, \ldots, K+1.\\
	\mbox{+ }  \xi_0 =\{p\}, \xi _{K+1} = \{q\}.\\
	\mbox{+ } \mathcal{F}_i \cap \mathcal{F}_{j} = \emptyset, \mbox{ for } i \neq j,\, i,j \in \{ 0,1, \ldots, K\}, \, \mathcal{F}=\bigcup_{i=0}^{K} \mathcal{F}_i.
\end{array} \right.
\end{equation}

\begin{figure}[htp]
%
\centering
\includegraphics[width=0.45\textwidth]{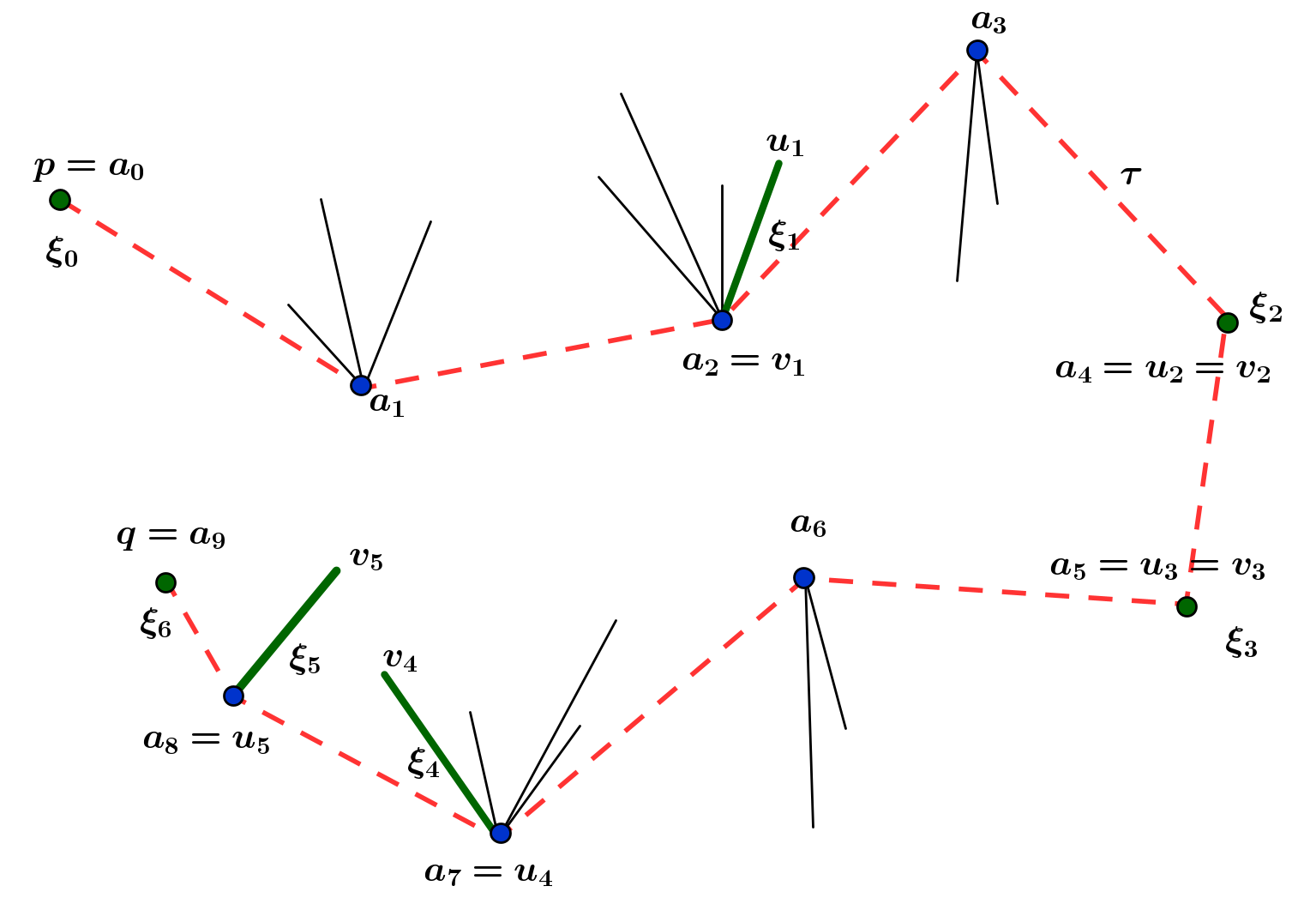}
\caption{$\mathcal{F}=\{a_0,a_1, \ldots,a_9\}$ is split  into sub-sequences: $\mathcal{F}_0 = \{a_0, a_1, a_2\},  \mathcal{F}_1 = \{a_3, a_4\}, \mathcal{F}_2 = \{a_5\}, \mathcal{F}_3 = \{a_6, a_7\}$ and $\mathcal{F}_4 = \{a_8\}$, where $N=8, K=4$. Cutting segments (bold  line segments):  $\xi_0 = \{a_0\}, \xi_1$ is the last line segment of  $\mathcal{F}_0$, $\xi_2 = \{a_4\}, \xi_3 = \{a_5\}, \, \xi_4$ and $\xi_5$ are the last line segments of  $\mathcal{F}_3$ and $\mathcal{F}_4$, respectively and $\xi_6 = \{a_9\}$}
\label{fig:partition}
\end{figure}

A partition given by  (\ref{eq:decomposition}) is illustrated in Figure~\ref{fig:partition}.
To be easy to describe the algorithm, we denote $\xi_i =[u_i, v_i]$, for all $i=0,1,\ldots, K+1$, in which  $v_i$ and $u_i$ are determined as follows:

\begin{align}
\label{eq:order-endpoint} 
\left\{ 	\begin{array}{lll}
	\mbox{  +  } \text{If } \xi_i \text{ has an endpoint lying on the right  of } \tau  \text{ when we traverse from } 
	p \text{ to } q,   \text{then } \\ v_i \text{ is this endpoint and } u_i \text{ is the remaining one.}\\
	\mbox{  +  }\text{Otherwise, } v_i \text{ is  the vertex of the bundle containing } \xi_i \text{ and } u_i \text{ is the remaining  } \\ \text{  endpoint.} 
\end{array} 
\right.
\end{align}
Here $u_0 = v_0= p, u_{K+1} =v_{K+1} = q$.

Firstly, we   initialize an ordered set of points by taking a point in each cutting segment. Two consecutive initial points are connected by the shortest path joining these points along the corresponding sub-sequence of bundles. The path received by combining these shortest  paths  is called the initial path of the algorithm (see Figure~\ref{fig:initial-path}). 
For convenience, these initial  points are chosen as $v_i$, for $i=0,1,\ldots,K+1$.
For each iteration step which is discussed carefully in next sections (\ref{sect:collinear_condition} and~\ref{sect:update}), we obtain a set of points $\{s_i | \, s_i \in \xi_i,  i=0,1,\ldots, K+1\}$ and a path $\gamma =\cup_{i=0}^{K} \text{SP}(s_i, s_{i+1})$, where $\text{SP}(s_i, s_{i+1})$ is the shortest path joining $s_i$ and $s_{i+1}$ along $\mathcal{F}_i$ and $s_0 = p, s_{K+1} = q$. 
Such a point $s_i$ is called a \textit{shooting point}. Then $\gamma$ is called \textit{the path formed by the ordered set of shooting points} $\{s_i |\, i=0,1,\ldots, K+1\}$ (see~\cite{An2017}).
%
Finding $\text{SP}(s_i, s_{i+1})$ is called by any known  algorithm for finding the shortest paths along a sequence of line segments. 
Throughout this paper, when  we say  $\text{SP}(x, y)$ without further explanation, it means the shortest path joining $x$ and $y$ along the corresponding sub-sequence of bundles of $\mathcal{F}$, where $x$ and $y$ belong to two distinct bundles of $\mathcal{F}$.  The corresponding sub-sequence of bundles of $\mathcal{F}$ consists of adjacent bundles between two bundles containing $x$ and $y$ in the given order of bundles in $\mathcal{F}$, and if $y$ belongs to the last segment of whose bundle, the bundle is also added into the corresponding sub-sequence.



\subsection{Factor (f2): establish and check the collinear condition}
\label{sect:collinear_condition}

Firstly, we present the so-called  \textit{collinear condition}. According to MMS,  $\text{SP}(s_i, s_{i+1})$ are found independently
along $\mathcal{F}_i (0 \le i\le K)$. If the collinear condition  holds at all shooting points, the shortest path is obtained by Proposition~\ref{prop:collinear-condition} in Section~\ref{sect:correctness}.

\begin{figure}[h]
\centering
\includegraphics[width=0.33\linewidth]{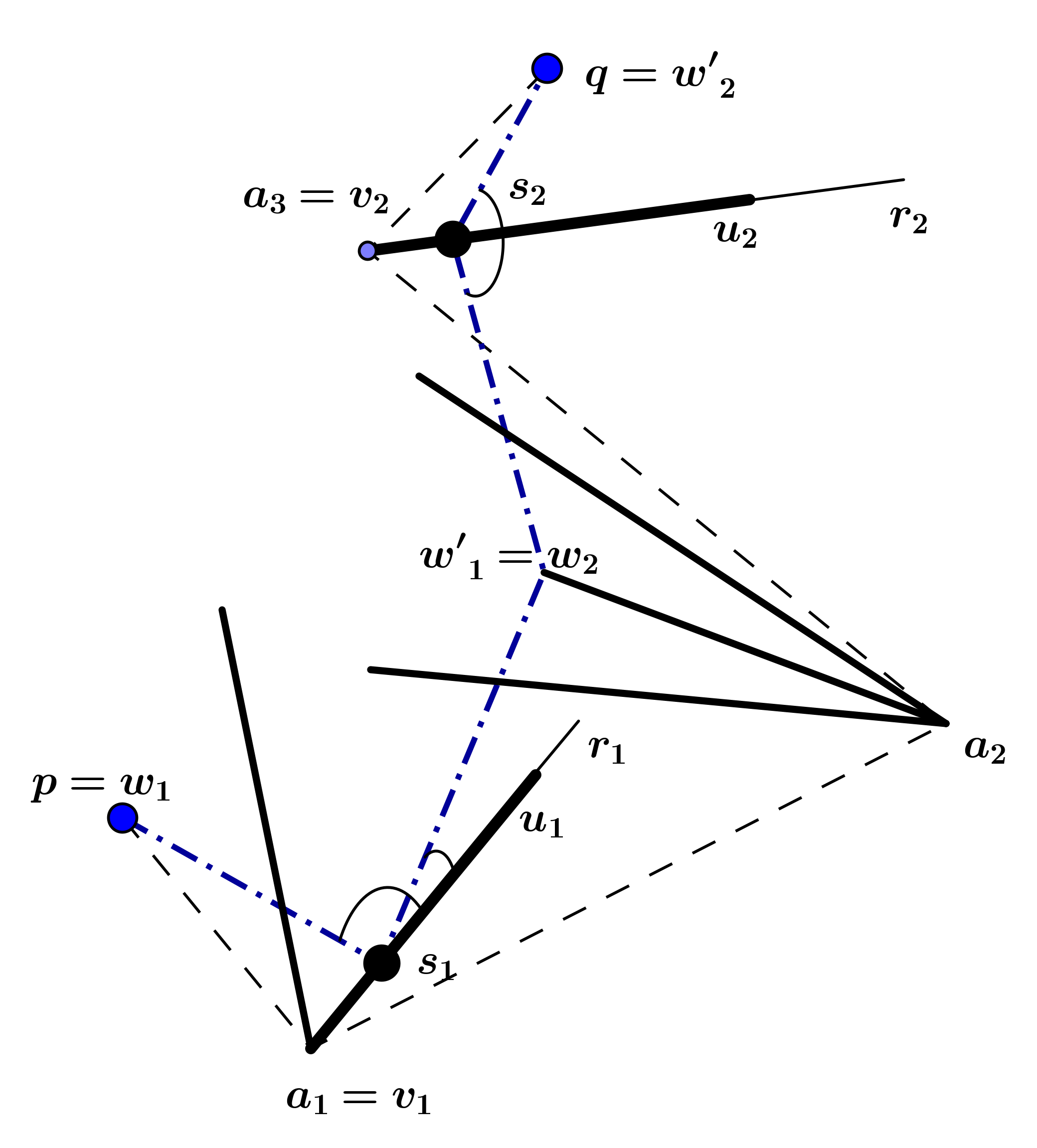}
\caption[Illustration of $\angle_{v_iu_i} \left(  \text{SP}( s_{i-1}, s_i), \text{SP}(s_i, s_{i+1}) \right) $]{Illustration of $\angle_{v_iu_i} \left(  \text{SP}( s_{i-1}, s_i), \text{SP}(s_i, s_{i+1}) \right) $, the angle at $s_1$ is the total of $\angle w_1s_1u_1$ and $\angle w'_1s_1u_1$, and the angle at $s_2$ is $\angle w_2s_2w'_2$}
\label{fig:2cases-same-side-or-not}
\end{figure}

\begin{figure}[htp]
\centering
\includegraphics[width=0.48\textwidth]{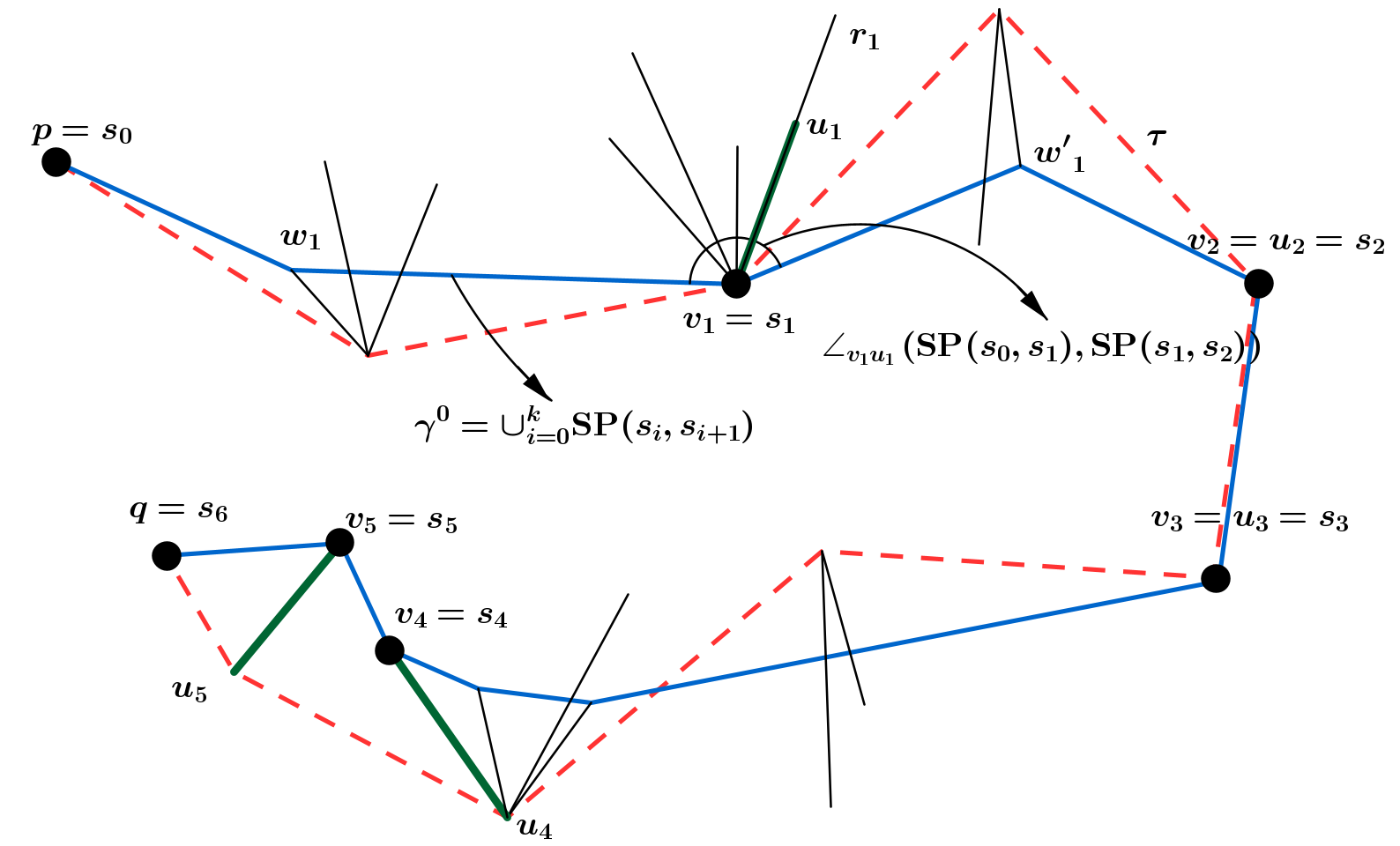}
\caption{Illustration of the set of initial shooting points $\{s_i = v_i, i=0,1, \ldots, K+1\}$ (black big dots), and the path (solid polyline) formed by  the set  and the angle $\angle_{v_1u_1} \left( \text{SP}( s_0, s_1), \text{SP}(s_1, s_2) \right)$ of the current path. }
\label{fig:initial-path}
\end{figure}

For $i=1,2,\ldots, K$, let $[w_i, s_i]$
and $[s_i,w'_i]$ be line segments of $\text{SP}(s_{i-1},s_i)$ and $\text{SP}(s_i,s_{i+1})$ respectively, sharing $s_i$ (see Figure~\ref{fig:initial-path}). 
If $u_i \neq v_i$, let $r_i$ be a point on the ray joining $v_i$ to $u_i$ such that $r_i$ is outside $[u_i,v_i]$.
The angle created by $\text{SP}(_{i-1}, s_i)$ and $\text{SP}(s_i, s_{i+1})$ with respect to the directed line $v_iu_i$, denoted by $\angle_{v_iu_i} \left(  \text{SP}( s_{i-1}, s_i), \text{SP}(s_i, s_{i+1}) \right) $, is defined to be the total angle $ \angle (\overrightarrow{w_i s_i}, \overrightarrow{s_i r_i}) + \angle (\overrightarrow{r_i s_i}, \overrightarrow{s_i w'_i})$, where we denote by $\angle (\overrightarrow{u}, \overrightarrow{v})$ the angle
between two nonzero vectors $\overrightarrow{u}$ and $\overrightarrow{v}$ in $\mathbb{R}^2$, which does not exceed $\pi$. 
 Figure~\ref{fig:2cases-same-side-or-not} illustrates two cases of $\angle_{v_iu_i} \left(  \text{SP}( s_{i-1}, s_i), \text{SP}(s_i, s_{i+1}) \right) $ depending on $w_i$ and $w'_i$ which are the same side to the line $u_iv_i$ or not. Herein, the polyline $\bigcup_{i=0}^{2} \text{SP}( s_{i}, s_{i+1})$ goes through $[u_2,v_2]$ and does not go through $[u_1,v_1]$, since the polyline reaches and reflects at $s_1$. 
%
%
For each $i=1,2,\ldots, K$, $s_i \in \xi_i$, the collinear condition states that:
\begin{itemize}
\item[(A1) ] Assume that $]u_i, v_i[$ is non-degenerate.
\par
\quad	- If $s_i \in ]u_i, v_i[$, we say that $s_i$   satisfies the collinear condition when the angle created by $\text{SP}( s_{i-1}, s_i)$ and $\text{SP}(s_i, s_{i+1})$ is equal to $\pi$ (i.e.,  $\angle_{v_iu_i} \left( \text{SP}( s_{i-1}, s_i), \text{SP}(s_i, s_{i+1})\right) =\pi$).
\par
\quad	- If $s_i =  u_i$, we say that $s_i$   satisfies  the collinear condition when the  angle created by $\text{SP}( s_{i-1}, s_i)$ and $\text{SP}(s_i, s_{i+1})$ is at least $\pi$  (i.e., $\angle_{v_iu_i} \left( \text{SP}( s_{i-1}, s_i), \text{SP}(s_i, s_{i+1})\right) \ge \pi$).
\par
\quad	- If $s_i = v_i$,  we say that $s_i$   satisfies  the collinear condition when the  angle created by $\text{SP}( s_{i-1}, s_i)$ and $\text{SP}(s_i, s_{i+1})$ is at most $\pi$  (i.e., $\angle_{v_iu_i} \left( \text{SP}( s_{i-1}, s_i), \text{SP}(s_i, s_{i+1})\right) \le \pi$).
\item[(A2)] 	Otherwise (i.e., $ u_i = v_i$), we say that $s_i$   satisfies the collinear condition.
\end{itemize}

\subsection{Factor (f3): update  shooting points}
\label{sect:update}
Suppose that at $j^{th}$-iteration step, we have a set of shooting points $\{s_i \vert \, s_i\in \xi_i=[u_i,v_i], i=0,1, \ldots, K+1\}$ and $\gamma =\cup_{i=0}^{K} \text{SP}(s_i, s_{i+1})$ is the path formed by the set of  shooting points, where $s_0 = p, s_{K+1} = q$. If Collinear Condition (A1-A2) holds at all shooting points, the algorithm stops. Otherwise,   shooting points are updated  such that the length of the path formed by the set of new shooting points is smaller than the length of the current path. 

\begin{figure}[htp]
\begin{subfigure}{.48\textwidth}
\raggedright
\includegraphics[scale=0.58]{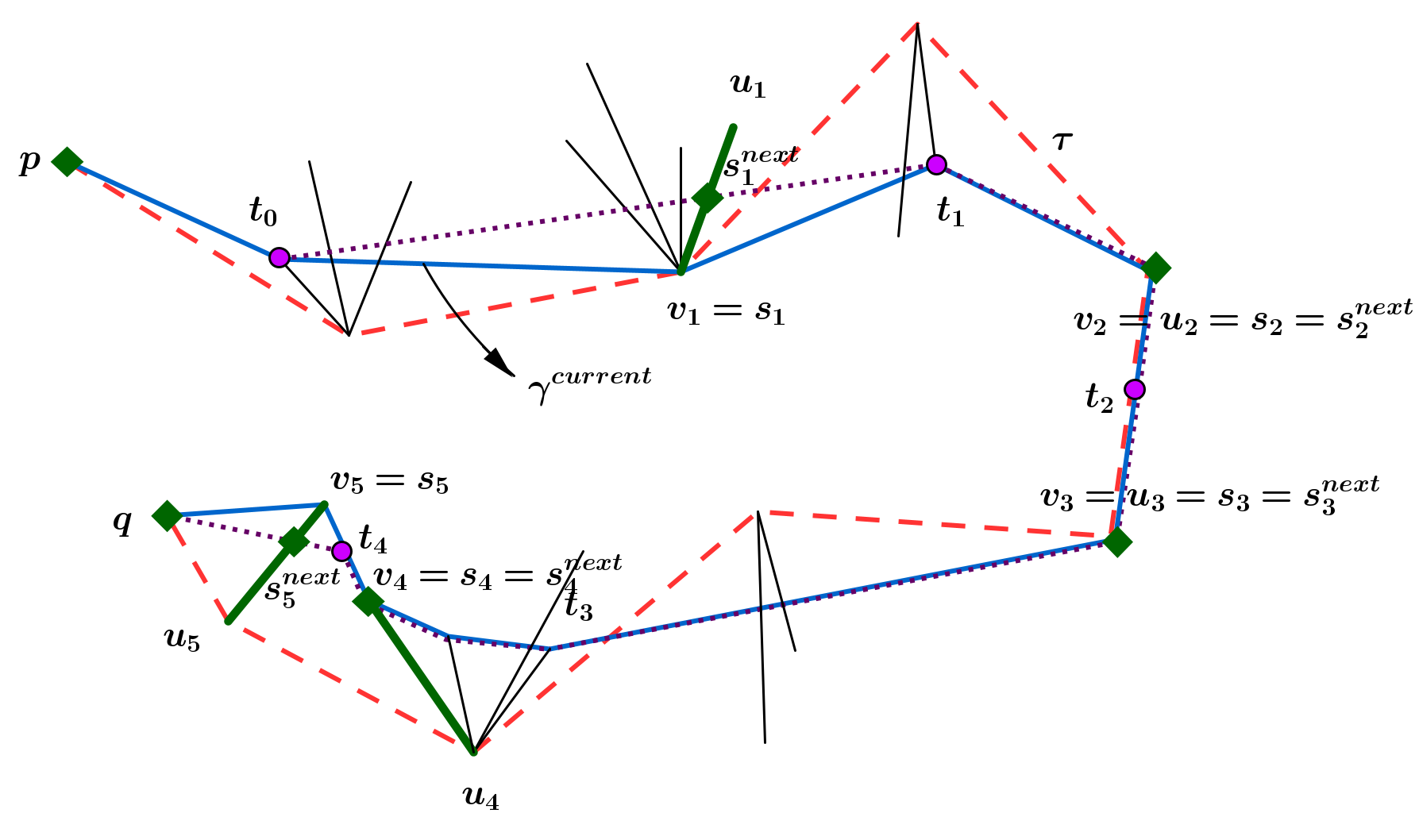}
\caption*{(i)}
\end{subfigure}
\begin{subfigure}{.52\textwidth}
\includegraphics[scale=0.58]{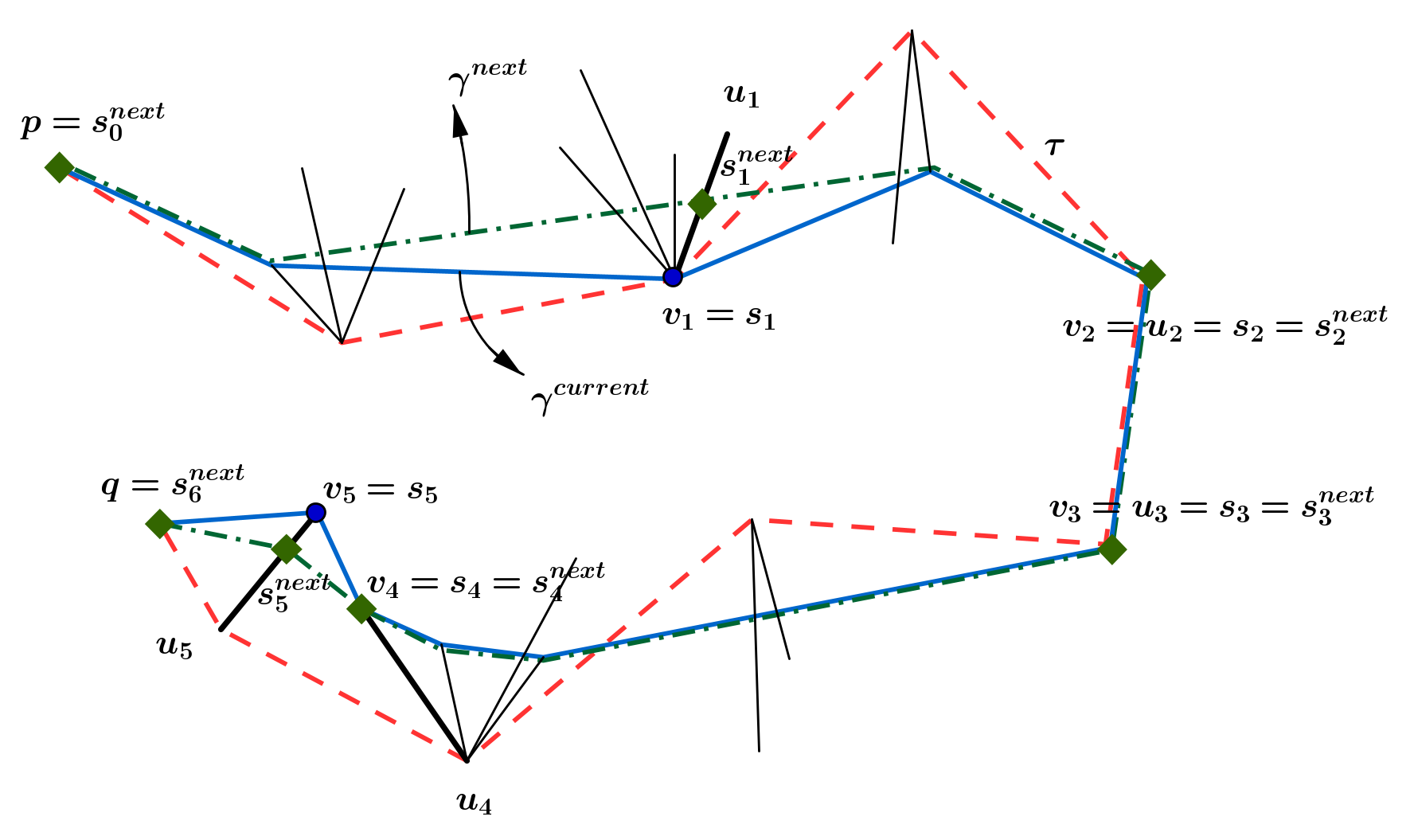}
\caption*{(ii)}
\end{subfigure}
\caption{(i): illustration of taking  points $\{t_i, i=0,1, \ldots, K\}$   and finding $s_i^{next}$ that is the intersection point of $\text{SP}(t_{i-1}, t_i)$ and $\xi_i$; (ii): illustration of  updating $\gamma^{current}$  (solid polyline) to $\gamma^{next}$ (dashed polyline)}
\label{fig:update-path}
\end{figure}

Assuming that Collinear Condition (A1-A2) does not hold at least one shooting point, we update  shooting points as follows:

\begin{itemize}
\item[(B)] 
For $i=1,2 \ldots, K$, take $t_{i-1} \in \text{SP}(s_{i-1}, s_i), t_i \in \text{SP}(s_i, s_{i+1})$ such that $t_{i-1}$ and $t_i$ are not identical to $s_{i-1}, s_i$ and $s_{i+1}$.
Let $s^{next}_i$ be one of intersection points of the shortest path $\text{SP}(t_{i-1}, t_i)$ 
and $[u_i, v_i]$.
We update  shooting point $s_i$ to $s^{next}_i$ (see Figure~\ref{fig:update-path}).
\end{itemize}

%

In particular, if $\text{SP}(s_i, s_{i+1})$ is a line segment, we choose $t_i$ as the midpoint of $\text{SP}(s_i, s_{i+1})$. Otherwise,  $t_i$ is chosen at one of endpoints  that is not identical to $s_i$ and $s_{i+1}$ of the polyline $\text{SP}(s_i, s_{i+1})$ (see Figure~\ref{fig:update-path}(i)). 
%
We can only update shooting points that do not hold the collinear condition and keep shooting points satisfying the collinear condition. Here our algorithm will update whole shooting points.
Because if Collinear Condition (A1-A2) holds at $s_i$, then $s_i$ and the new update shooting point of $s_i$ are the same  due to Proposition~\ref{prop:two-stop-condition} (in Section~\ref{sect:correctness}).
For example in Figure~\ref{fig:update-path}(i), Collinear Condition (A1-A2) holds at shooting points $s_2, s_3$, and $s_4$ for $\gamma^{current}$. The update given by (B) shows that $s_2^{next}=s_2, s_3^{next}=s_3$ and $s_4^{next}=s_4$. 
%


\subsection{Pseudo-code of the proposed algorithm}
\label{sect:main-algorithm}
\begin{algorithm}[H]
\caption{{Find approximately the shortest path joining two points along a sequence of bundles of line segments}}
\label{alg:main}
\begin{algorithmic}[1]
\Require Two points $p$ and $q$, a sequence $\mathcal{F}$ of bundles of line segments.
\Ensure An approximate shortest path joining $p$ and $q$ along $\mathcal{F}$.
%
\State Split $\mathcal{F}$ into $\mathcal{F}_i$ satisfying~(\ref{eq:decomposition}) 
by cutting segments $[u_i,v_i]$, where the order of $u_i$ and $v_i$ is shown in~(\ref{eq:order-endpoint}), for $i=0, 1,\ldots,K+1.$
\label{step1}
\State Take an ordered set $\{s_i\}_{i=0}^{K+1}$ of initial shooting points  consisting of $p, v_i (i=1,2,\ldots,K)$ and $q$. Let $\gamma^{current}$ be the path formed by $\{s_i\}_{i=0}^{K+1}$.
\label{step2}
\State $flag \leftarrow true$, $s_i^{next} \leftarrow s_i$ and $\gamma^{next}$ is the path formed by $\{s_i^{next}\}_{i=0}^{K+1}$.
\label{step3}
\State Call Procedure {\sc Collinear\_Update}($\gamma^{current}, \gamma^{next}, flag$).
\label{iteration}
\If{Collinear Condition (A1-A2) holds at all shooting points, i.e., $flag= true$}		
\State
{\bf stop} and \Return $\gamma^{next}$. 
\Comment{{\it \color{blue}by Prop.~\ref{prop:collinear-condition}, the shortest path is obtained}}
\Else 	\;	 
$\gamma^{current} \leftarrow \gamma^{next}$ and  
\textbf{go to } step~\ref{iteration}.
\EndIf
%
\State	\Return $\gamma^{next}$.

\end{algorithmic}
\end{algorithm}

If Collinear Condition (A1-A2) holds at all shooting points, Algorithm~\ref{alg:main} stops after a finite number of iteration steps and the shortest path is obtained by Proposition~\ref{prop:collinear-condition}. Otherwise, \break Procedure {\sc Collinear\_Update}($\gamma^{current}, \gamma^{next}, flag$) performs updating shooting points to get a better path $\gamma^{next}$ by Proposition~\ref{prop:update_path}.

\newfloat{procedure}{htbp}{loa}
\floatname{procedure}{Procedure}

\begin{procedure}[h!]
\begin{algorithmic}[1]
\caption*{\textbf{Procedure} \sc Collinear\_Update($\gamma^{current}, \gamma^{next}, flag$)} \label{alg:update}
\Require $\gamma^{current}$ is the union of the shortest paths joining pairs of shooting points $s_i$, i.e., \break $\gamma^{current}=\cup_{i=0}^K \hbox{SP}(s_i,s_{i+1})$
\Ensure Determine if  $flag=true $  or $flag = false $ and the set of new shooting points 
such that $l(\gamma^{next}) \le l(\gamma^{current})$, where $\gamma^{next}$ is the union of the shortest paths joining pairs of new shooting points $s_i^{next}$
\State  $flag \leftarrow true$
\For {$i=0,1, \ldots, K$}
\State Take points $t_i$s which are defined
in (B)   such that \par
\quad	if $\text{SP}(s_i, s_{i+1})$ is a line segment, $t_i$ is the midpoint of $\text{SP}(s_i, s_{i+1})$, \par
\quad	otherwise,  $t_i$ is  one of endpoints  that is not identical to $s_i$ and $s_{i+1}$.
\If { $s_i \in \, ]v_i, u_i[$} 
\If {$\angle_{v_iu_i} \left( \text{SP}( s_{i-1}, s_i), \text{SP}(s_i, s_{i+1}) \right) =\pi$} 
\Comment{{\it \color{blue}  Collinear Condition {\rm (A1-A2) is checked}}}
\State Set $s_i^{next} \leftarrow s_i$
\Else \ call $\text{SP}(t_{i-1}, t_i)$ 
\State Set $s_i^{next}$ be the intersection point of  $\text{SP}(t_{i-1}, t_i)$ and $[u_i, v_i]$
\Comment{{\it \color{blue} due to {\rm (B)}}}
\label{prcRobot:step8}
\State $flag \leftarrow false$
\EndIf 
\ElsIf{$s_i = v_i$}
\If {$\angle_{v_iu_i} \left( \text{SP}( s_{i-1}, s_i), \text{SP}(s_i, s_{i+1}) \right) \ge \pi$} 
\Comment{{\it \color{blue} Collinear Condition {\rm (A1-A2) is checked}}}
\State Set $s_i^{next} \leftarrow s_i$
\Else \ call $\text{SP}(t_{i-1}, t_i)$ 
\State Set $s_i^{next}$ be the intersection point of $\text{SP}(t_{i-1}, t_i)$ and $[u_i, v_i]$
\Comment{{\it \color{blue} due to {\rm (B)}}}
\label{prcRobot:step14}
\State $flag \leftarrow false$
\EndIf 
%
\ElsIf{$s_i = u_i$}
\If {$\angle_{v_iu_i} \left( \text{SP}( s_{i-1}, s_i), \text{SP}(s_i, s_{i+1}) \right) \le \pi$} 
\Comment{{\it \color{blue} Collinear Condition {\rm (A1-A2) is checked}}}
\State Set $s_i^{next} \leftarrow s_i$
\Else \ call $\text{SP}(t_{i-1}, t_i)$ 
\State Set $s_i^{next}$ be the intersection point of $\text{SP}(t_{i-1}, t_i)$ and $[u_i, v_i]$
\Comment{{\it \color{blue} due to {\rm (B)}}}
\label{prcRobot:step20}
\State $flag \leftarrow false$
\EndIf 
\EndIf
\EndFor
\State $s_0^{next} \leftarrow p, s_{K+1}^{next} \leftarrow q$
\end{algorithmic}
\end{procedure}
Theoretically, it is necessary to have an available algorithm for finding the shortest path along a sequence of bundles of line segments to solve sub-problems of the MMS-based algorithm after partitioning. However, a general algorithm for the SP problem for arbitrary line segments is not yet available. 
Nevertheless, if we partition the sequence of $N+2$ bundles into $N$ sub-sequences of bundles by taking one cutting segment at each bundle except $p$ and $q$ (i.e., $K=N$),
then each sub-problem of the MMS-based algorithm can be dealt with completely by the funnel technique. As each sub-problem only consists of one bundle, it becomes the problem for finding the shortest path along a sequence of diagonals, which are line segments of the bundle, inside a simple\textsf{} polygon being the convex hull of the bundle. Then it can be dealt with by the funnel technique. 
\section{The correctness  of Algorithm~\ref{alg:main}}
\label{sect:correctness}

Let $\gamma^*$ be the shortest path joining $p$ and $q$ along $\mathcal{F}$ (i.e., $\gamma^* =\text{SP}(p,q)$). Applying Corollary 4.3 in~\cite{HaiAnHuyen2019}, we conclude the remark below:

\begin{remark}
\label{rem:optimal_solution}	
Suppose that the shortest path $\gamma^*$  intersects with $\xi_i$ at $s_i$, for $i=1,2, \ldots, K$.  Then Collinear Condition (A1-A2) holds at all shooting points $s_i$,  for all $i=1,2,\ldots, K$.
\end{remark}


\begin{proposition}
\label{prop:collinear-condition}
Let $\gamma$ be a path formed by a set of shooting points, which joins $p$ and $q$ along $\mathcal{F}$.
If Collinear Condition (A1-A2) holds at all  shooting points, then $\gamma$ is indeed the shortest path $\gamma^*$.
\end{proposition}


\begin{proposition}
\label{prop:two-stop-condition}
Collinear Condition (A1-A2) holds  at  $s_i$ if and only if $\text{SP}(t_{i-1}, t_i)\cap [u_i, v_i] = s_i$, where $t_{i-1}$ and $t_i$ are defined
in (B).  
\end{proposition}

\begin{proposition}
\label{prop:update_path} The update of Algorithm~\ref{alg:main} by factor (f3), which is \break Procedure {\sc  Collinear\_Update}($\gamma^{current}, \gamma^{next}, flag$), gives 
\begin{align*}
l\left(\gamma^{next} \right) \le l\left( \gamma^{current}\right).
\end{align*}

Thus the  sequence of lengths of paths  obtained by Algorithm~\ref{alg:main} is convergent. 	If Collinear Condition (A1-A2) does not hold at some shooting point, then the inequality
above is strict.
\end{proposition}

\begin{theorem}
\label{theo:global solution}
Assume that every two distinct bundles of $\mathcal{F}$ does not intersect each other,
and every polyline joining $p$ and $q$ along $\mathcal{F}$, whose endpoints except $p$ and $q$ only lie on line segments of $\mathcal{F}$, goes through line segments of $\mathcal{F}$.
Then Algorithm~\ref{alg:main} is convergent. It means that if the algorithm stops after a finite number of iteration steps (i.e.,  Collinear Condition {\rm (A1-A2)} holds at all shooting points), we obtain the shortest path $\gamma^*$, otherwise, 
sequence of paths $\{\gamma^{j}\}$ converges to $\gamma^*$ by Hausdorff distance. 
\end{theorem}
 To discuss the complexity of Algorithm~\ref{alg:main}, assume that the partition of $\mathcal{F}$ results in the number of bundles of each sub-sequence of $\mathcal{F}$ not exceeding a constant $c$. For instance, we can partition uniformly in which the number of $K$ cutting segments except $p$ and $q$ can be taken as the floor of $N/c$ (i.e., $K = \lfloor N/c \rfloor$), where $1 \le c \le N$. Based on this assumption, we have the following theorem: 
\begin{proposition}
\label{theo:complexity}
Algorithm~\ref{alg:main} runs in $O(mNJ)$ time, where $N$ is the number of bundles of $\mathcal {F}$, $J$ is the number of line segments of $\mathcal {F}$, and $m$ is the  number of iterations to get the required path joining $p$ and $q$.
\end{proposition}

The proofs of Proposition~\ref{prop:collinear-condition}-\ref{theo:complexity}, and  Theorem~\ref{theo:global solution} are  given in the Appendices.

%
%
%

 \section{An application in robotics} 
\label{sect:refer_problem}

This section presents the autonomous robot problem  originating from the problem of finding the shortest path along a sequence of bundles of line segments.

\subsection{The path planning problem for autonomous robots in unknown environments in 2D}
\label{sect:autonomous_robot}

\begin{figure}[htp]
\begin{subfigure}{.48\textwidth}
\raggedright
\includegraphics[scale=0.56]{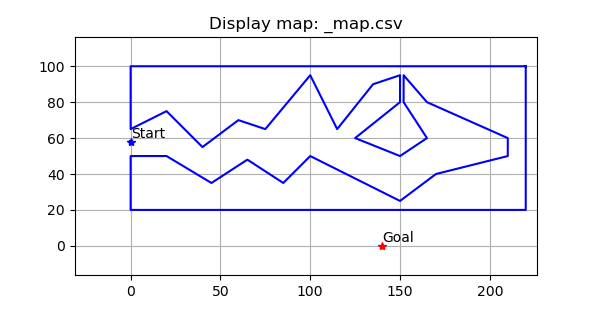}
\caption*{(i)}
\end{subfigure}
\begin{subfigure}{.5\textwidth}
\includegraphics[scale=0.58]{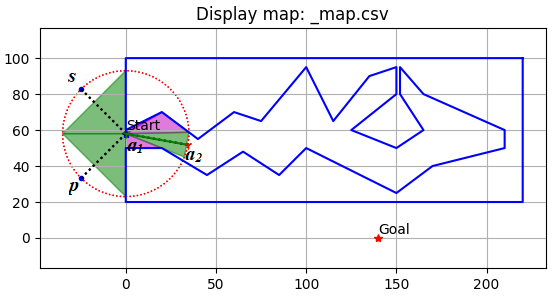}
\caption*{(ii)}
\end{subfigure}
\centering
\caption{ A robot in an unknown environment starts at $a_1$.
The intersection between $\overline{B}(a_1,r)$ and collision-free space includes: the sectors of $\overline{B}(a_1,r)$ corresponding to the green triangles  illustrate illustrating  open sights where the robot can move safely whereas  pink triangles demonstrate closed sights where the robot can meet dead-end}
\label{fig:vision-range}
\end{figure}

Recall that an autonomous robot moving in an unknown environment in 2D has a vision that is a circle centered at its current position, say $a$, with a certain radius $r$. Let us denote  by $\overline{B}(a,r)$ the robot's vision range at $a$. The problem is to find a path from starting point to the goal avoiding polygonal obstacles. The robot does not have full map information except for the coordinates of starting and ending points. 
The obstacles  are represented by the regions bounded by the union of polylines, such as the polygon shown in Figure~\ref{fig:vision-range}. The path of the robot has to 
avoid the obstacles.
%
Let us consider the intersection between $\overline{B}(a,r)$  and \textit{collision-free space} which is  the complement of the union of the interiors of the obstacles. As shown in Figure~\ref{fig:vision-range}, the sectors of $\overline{B}(a,r)$ corresponding to the green triangles  illustrate \textit{open sights}   that robot can move in the next steps whereas the pink triangles demonstrate \textit{closed sights} that give   collision warnings. Each open sight has the so-called \textit{open point} which is the midpoint of each arc with respect to the sector determining one open sight. 
Clearly, moving to the directions from the center $a$ to a  point in closed sights, the robot will hit a dead-end. Then at the next step, it must find one open point to move to. 
%
At each position, the robot has many open points to go and it must determine which one is selected. \textit{Ranking} each open point   estimates the possibility of leading to the goal from that direction. Each open point and its rank is identified by a predetermined rule which is established carefully in local path planning in~\cite{BinhAnHoai2021}. In particular, there are two factors including distance (between the open point and the  goal, denoted by $d(t)$) and angle (forming between $a$ to the  open point and $a$ to the  goal, denoted by $a(t)$). Then
\begin{align}
\label{eq:rank}
rank (t) =\left\{
\begin{array}{ll}
\dfrac{\alpha}{d(t)} + \dfrac{\beta}{a(t)} & \text{ if } d(t) \ne 0 \text{ and } a(t) \ne 0\\
+\infty & \text{ if } d(t) =0 \text{ or } a(t) = 0
\end{array}
\right. , 
\end{align}
where $\alpha$ and $\beta$ are set based on how important the angle and distance are. In implementation, we take  $\alpha =\beta =1$, 
where the  open point with  the highest rank would have first priority. Then we need to store a graph $G$ and a set  $OP$.  Here the graph $G$ consists of nodes which are centers and edges which are radial line segments joining nodes. The set $OP$  consists of open points and their rank. Furthermore, in storing efficiently nodes of the graph $G$, each angle of the sector determining each open sight does not exceed $\pi/3$, and each open point does not belong to any previous open or closed sights.
%
%


\begin{figure}[htp]
\begin{subfigure}{.48\textwidth}
\raggedright
\includegraphics[scale=0.58]{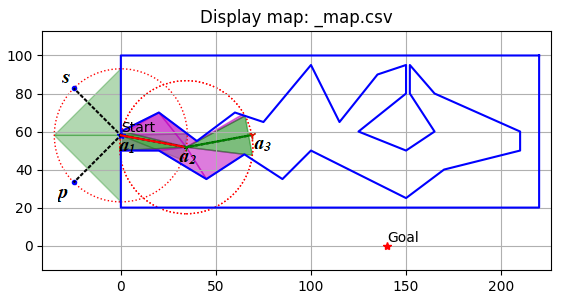}
\caption*{(i)}
\end{subfigure}
\begin{subfigure}{.52\textwidth}
\includegraphics[scale=0.58]{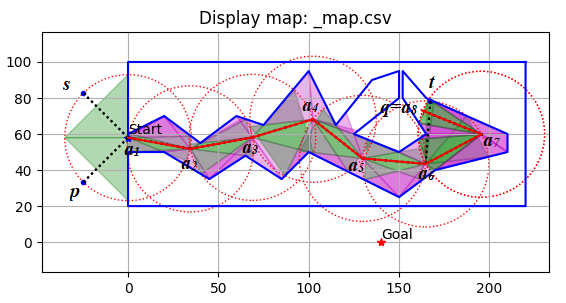}
\caption*{(ii)}
\end{subfigure}
\caption{After moving to the open point $a_2$,  the intersection between $\overline{B}(a_2,r)$ and collision-free space  determines open sights, open points. The robot moves to the next open point $a_3$ and it is continued until reaching $a_8$.}
\label{fig:moving-range}
\end{figure}

\begin{figure}[htp]
\begin{subfigure}{.48\textwidth}
\raggedright
\includegraphics[scale=0.5]{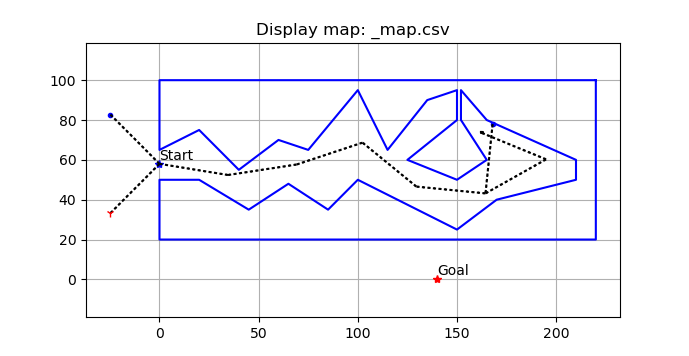}
\caption*{(i)}
\end{subfigure}
\begin{subfigure}{.52\textwidth}
\includegraphics[scale=0.54]{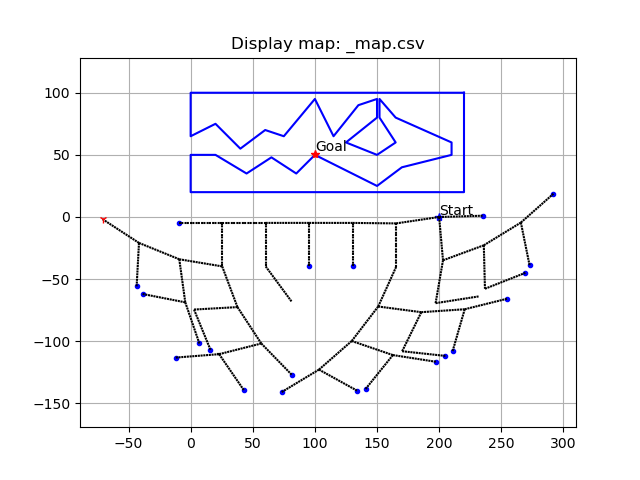}
\caption*{(ii)}
\end{subfigure}
\caption{Illustrating the graphs $G$ in some cases; (i): the graph $G$ at the step of finding the shortest path joining $q$ and $p$ in Example 1 and (ii): the graph $G$ in some step of another case with starting and ending point given in the figure.}
\label{fig:graph}
\end{figure}

\textbf{Example 1.} Assume that the starting point and the goal are shown in Figure~\ref{fig:vision-range}.  Initial $G= \emptyset, OP =\emptyset$. At $a_1$, the robot has three open points $p, s$, and $a_2$ which do not belong to any previous sight. Then they are added into $OP$. Then $a_1$ and endpoints of line segments representing closed and open sights at $a_1$ are added into $G$.
When	$a_2$ is an open point having  the highest rank, the robot keeps moving to $a_2$, then $a_2$ is added into $P$. At $a_2$, there are two open points $a_1$ and $a_3$, but $a_1$ is contained in a previous open sight. Therefore only $a_3$ is added into $OP$ and $a_2$ is removed from $OP$. After considering $a_2$, endpoints of line segments representing closed sights and the open sight corresponding to $a_3$ are added into $G$. The process of finding the path  from $a_1$ to $a_8$ is shown in Figures~\ref{fig:vision-range} and~\ref{fig:moving-range}. 
%
%
%
After moving to $a_8$, $OP$ consists of $s$, $p$ and $t$.
The robot now recognizes that the rank of the open point $p$   is  highest, it then  goes to $p$. 
Let $q=a_8$. Since  $p$ does not belong to the robot's vision at $q$, i.e., $p \notin \overline{B}(q,r)$,  a graph search-based  algorithm  is used to find  the shortest path  joining $q$ and $p$ in $G$.   The graph $G$ at this step is shown in Figure~\ref{fig:graph}(i). Note that in many cases, the graph $G$ is more complex, such as a case illustrated in Figure~\ref{fig:graph}(ii).
%
Such a path
is the polyline  joining $q,a_7, a_6, \ldots, a_1,p$.
A popular approach of path planning algorithms for robots is to use   shortest paths in  graphs  for  going from $p$ to $q$.
However, in order to make an efficient motion, we will find the shortest path joining $p$ and $q$ along a sequence of bundles of line segments, where the construction of these bundles  attaching with the polyline joining $q,a_7, a_6, \ldots, a_1,p$ are described in detail in Section~\ref{sect:SP_problem}.
Going back by the shorter path than the polyline joining $q,a_7, a_6, \ldots, a_1,p$ makes it time-saving for actual robots. 

\subsection{The SP problem for bundles of line segments as a phase of the path planning problem} 
\label{sect:SP_problem}

\begin{figure}[htp]
\begin{subfigure}{.46\textwidth}
\raggedright
\includegraphics[width=1.06\linewidth]{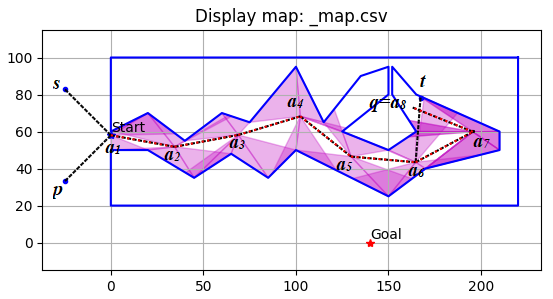}
\caption*{(i)}
\end{subfigure}
\begin{subfigure}{.5\textwidth}
\vspace*{0.8cm}
\includegraphics[width=1.1\linewidth]{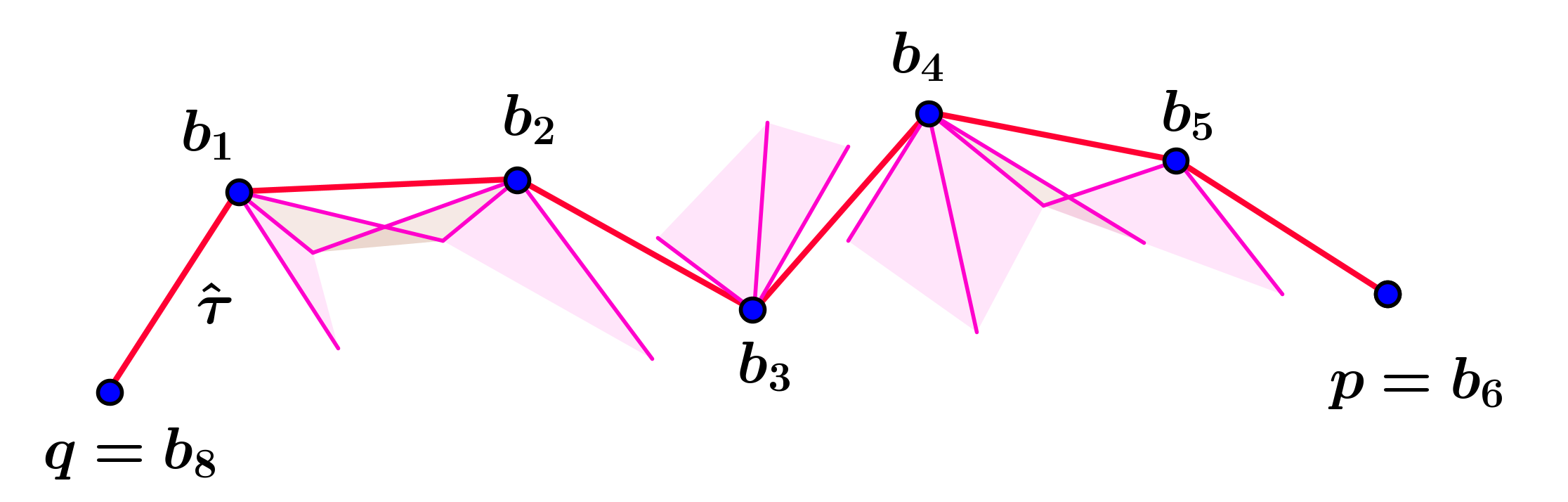}
\vspace*{0.5cm}
\caption*{(ii)}
\end{subfigure}
\caption{Initial sequences of bundles of line segments are reduced to the sequence of bundles of line segments in which the angle of the sector of $\overline{B}(a_i,r)$ containing the bundle $a_i$ does not exceed $\pi$, for $i = 1,2, \ldots,N$}
\label{fig:auto-robot}
\end{figure}

This section  shows the necessity of solving Problem (P) in path-planing for  autonomous robots in unknown environments.
Let $\hat{\tau}$ be the shortest path joining $p$ and $q$ in the graph $G$, where $q$ is the current position of the robot and $p \notin \overline{B}(q,r)$. Assume that $\hat{\tau}$ is the polyline joining $a_0, a_1, \ldots, a_{N+1}$, where $a_i$s are centers that the robot traversed, $a_0 =p,  a_{N+1}=q$.
Then  there are always more than two vertices of bundles attaching with $\hat{\tau}$. 
According to Problem (P$^*$), our object is to find a path (or shortest path) $\gamma$ joining $p$ and $q$ avoiding the obstacles, which is shorter than $\hat{\tau}$ (i.e., $l(\gamma) < l(\hat{\tau})$). 
The paths   determined by  (P$^*$) go through  a sequence 
of bundles of line segments attaching with $\hat{\tau}$. These bundles are obtained from a set of  triangles representing  sights. 
At  $a_i$, one of three cases occurs: there are two opposite non-degenerate bundles; there is only one non-degenerate bundle or there is no non-degenerate bundle, see Figure~\ref{fig:auto-robot}(i). Thus there  can be up to $2^{N-1}$ sequences of bundles of line segments attaching with $\hat{\tau}$. 
%
To reduce the number of computational cases,
at $a_i$, we only consider the bundle contained in the minor sector of  $\overline{B}(a_i,r)$. Note that for $i = 1,2, \ldots,N$, there are two sectors  enclosed by two line segments $[a_{i},a_{i-1}]$ and $[a_{i},a_{i+1}]$ and arcs of $\overline{B}(a_i,r)$.
%
Let us denote the sequence of bundles of line segments $\{a_0, a_1, \ldots, a_{N+1} \}$ by $\mathcal{F}^*$, in which the angle of the sector of $\overline{B}(a_i,r)$ containing the bundle $a_i$ does not exceed $\pi$, for $i = 1,2, \ldots,N$.
If the angle of the  sector of  $\overline{B}(a_i,r)$ containing the bundle $a_i$ is less than $\pi$, we can shorten $\hat{\tau}$ at the bundle's vertex. Here the shortening technique can be seen in
\cite{An2017}. Therefore, we get the following remark:
\begin{remark}
\label{obser:reduce-bruches}
If there is at least one non-degenerate bundle $a_i$ in which the angle of the  sector of  $\overline{B}(a_i,r)$ containing the bundle $a_i$ is  less than $\pi$, then there exists a path joining $p$ and $q$ along $\mathcal{F}^*$  which is shorter than $\hat{\tau}$.
\end{remark}
According to Remark~\ref{obser:reduce-bruches},  bundles contained in the  sector of  $\overline{B}(a_i,r)$ having the angle exceeds $\pi$ are removed to reduce computational load,
because our object is to find a path shorter than $\hat{\tau}$,
see Figure~\ref{fig:auto-robot}(ii). 
%

%

%


Now, if there are two distinct bundles of $\mathcal{F}^*$ intersecting each other, then it is not possible to use  the technique of Li and Klette~\cite{Li2011} for   finding approximately the shortest path joining two points along a sequence of  line segments having common points
even when we use modifications as established in Section~\ref{sect:numerical_results}. Besides, Algorithm~\ref{alg:main}  provides a sequence of paths whose lengths are decreasing, but the sequence its self is not convergent (Theorem~\ref{theo:global solution} is not true). Therefore the original sequence $\mathcal{F}^*$ of bundles must be preprocessed  to ensure that two distinct bundles do not intersect each other. 
We preprocess $\mathcal{F}^*$ by  Procedure~{\sc Perprocessing}$(\mathcal{F}^*)$ as follows:

\begin{algorithm}[H] 
\begin{algorithmic}[1]
\Procedure{Preprocessing}{$\mathcal{F}^*$}
\label{procedure:preprocessing}
\Require A  sequence of bundles of line segments $\mathcal{F}^* = \{ a_0, a_1, \ldots, a_{N+1}\}$ attaching with  $\hat{\tau}$; the angle of the sector of $\overline{B}(a_i,r)$ containing the bundle $a_i$ does not exceed $\pi$, for $i=1,2,\ldots, N$.
\Ensure A sequence of bundles $\hat{\mathcal{F}}$ of line segments.  
\For{$i=1$ to $N$}
%
\State Construct $\overline{B}(a_i,r_0)$, where $r_0 \le \frac{1}{2}\min\limits_{\substack{ 0 \le j,k \le N \\ j\neq k}} \Vert a_j - a_k\Vert$.
\label{step:set_radius}
\For{\textbf{each} line segment $[a_i, b_j]$ of the bundle $a_i$}
\Comment{{\it \color{blue} $a_i$ consists of $[a_i,b_1], [a_i,b_2], \ldots, [a_i,b_m]$}}
\If{$\Vert a_i-b_j \Vert > r_0$}
\Comment{{\it \color{blue} $b_j$ is outside of $\overline{B}(a_i,r_0)$}}
\State $b_j \leftarrow a_i+r_0\dfrac{a_i-b_j}{\Vert a_i-b_j \Vert }$
\Comment{{\it \color{blue} $b_j$ is replaced  by  $[a_i, b_j]\cap \text{bd}\overline{B}(a_i,r_0),r_0)$ }}
\EndIf
\EndFor

\EndFor
\EndProcedure
\end{algorithmic}
\end{algorithm}

\begin{figure}[ht!]
\begin{subfigure}[t]{.5\textwidth}
\centering
\includegraphics[scale=0.38]{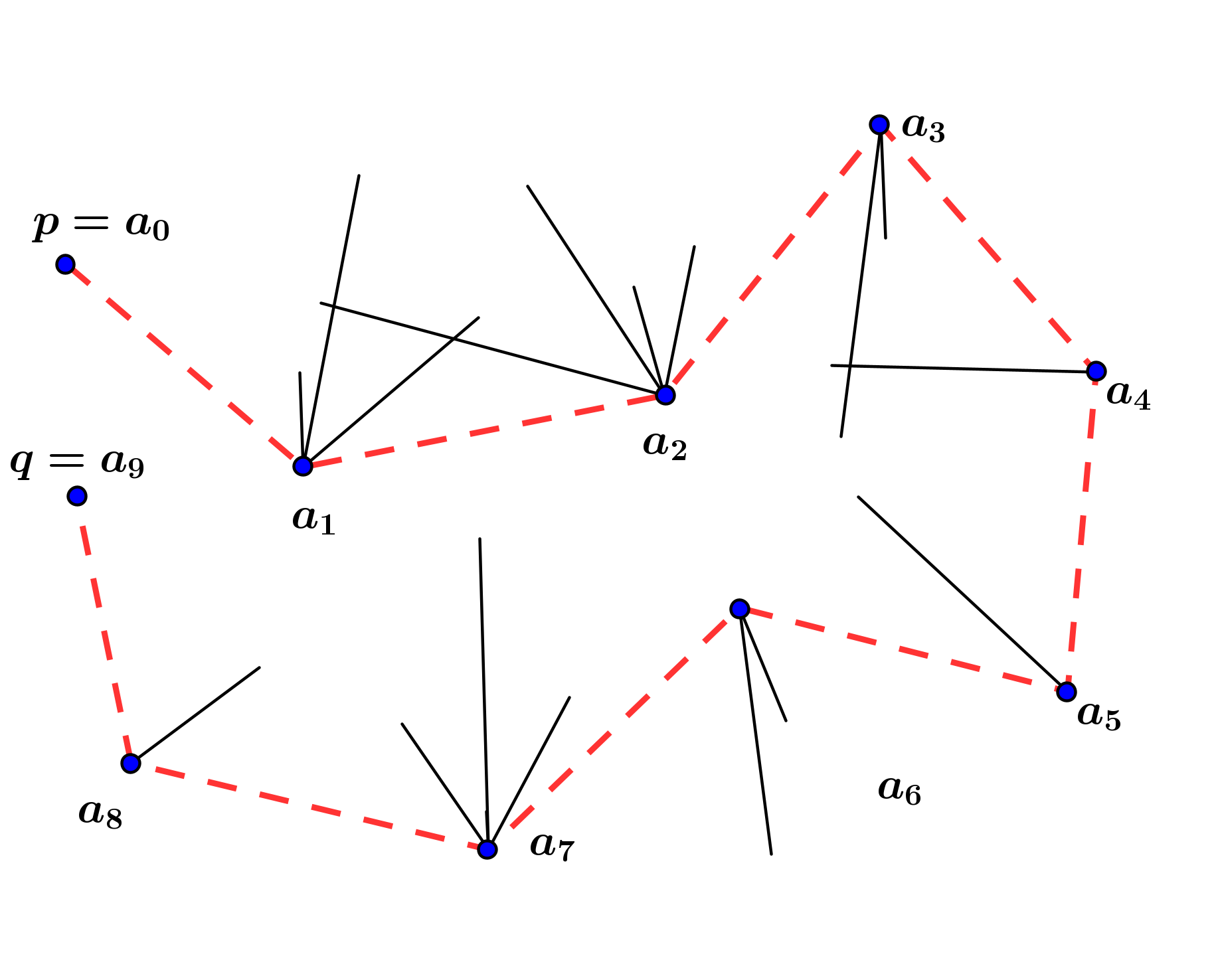}
\caption*{(i)}
\end{subfigure}
\begin{subfigure}[t]{.5\textwidth}
\centering
\includegraphics[scale=0.38]{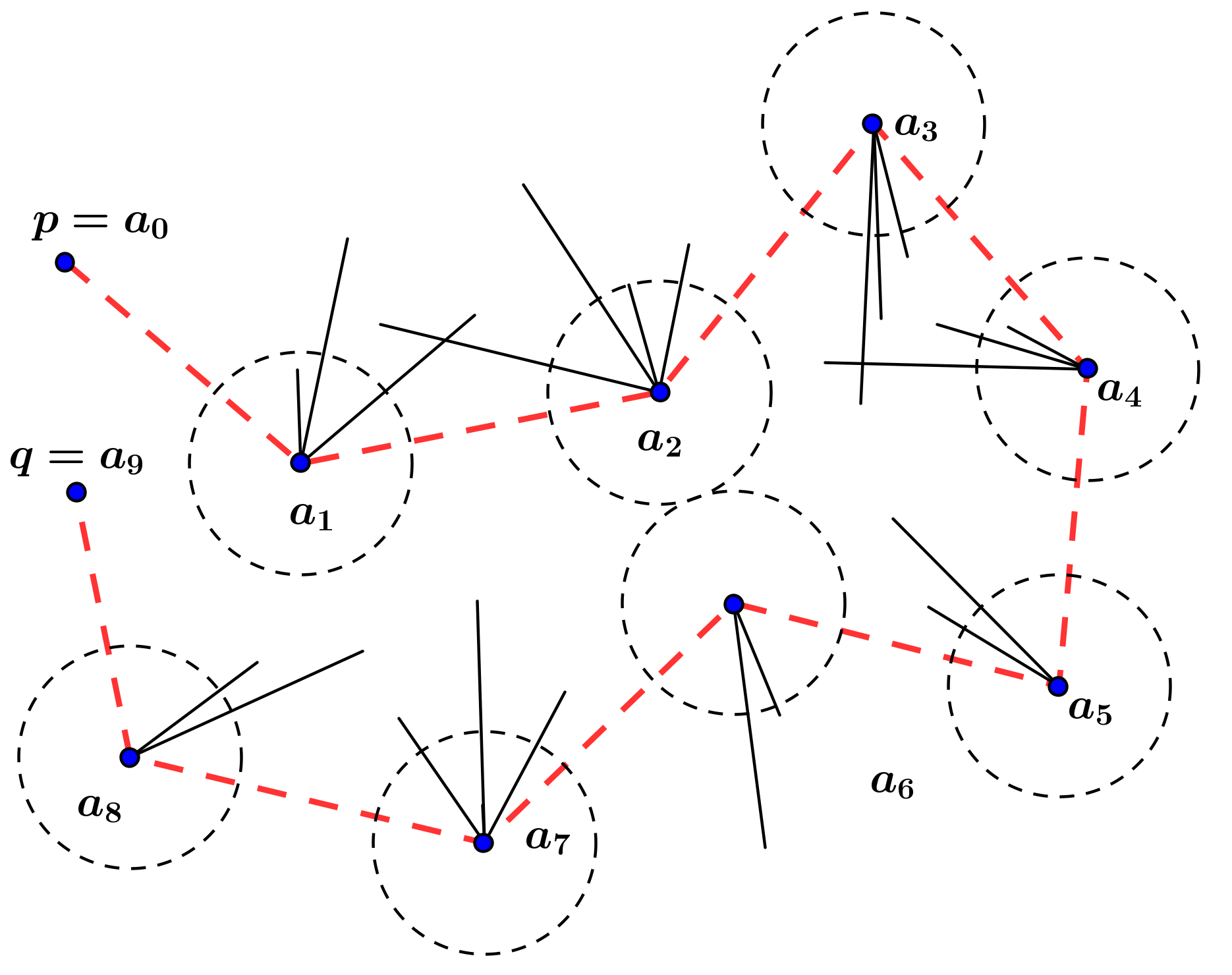}
\caption*{(ii)}
\end{subfigure}
\caption{Illustration of preprocessing $\mathcal{F}^*$ to obtain $\hat{\mathcal{F}}$ according to Procedure~{\sc Preprocessing}$(\mathcal{F}^*)$}
\label{fig:preprocessing}
\end{figure}


\begin{proposition}
\label{obser:reduce_cross} 
The obtained sequence of  bundles 
$\hat{\mathcal{F}}  =${\sc Perprocessing}$(\mathcal{F}^*)$ satisfies:
\begin{itemize}
\item[(a)]  Two distinct bundles of $\hat{\mathcal{F}}$ do not intersect each other.
\item[(b)] The shortest path joining $p$ and $q$ along  $\hat{\mathcal{F}}$ avoids the obstacles.
\end{itemize}

\end{proposition}

The proof of Proposition~\ref{obser:reduce_cross} is given in the Appendices.
Furthermore,  each bundle $a_i$ of $\hat{\mathcal{F}}$  belongs to the sector of $\overline{B}(a_i,r)$ whose angle does not exceed $\pi$, for $i = 1,2, \ldots,N$.
%
%
Thus, instead of solving Problem (P$^*$), we deal with   (P) which  is our object in this paper.


The following path planning algorithm (Algorithm~\ref{alg:path_planning}) for autonomous robots   is modified from An et al.'s algorithm in~\cite{BinhAnHoai2021}. Whenever the robot moves from $q$ to a position marked in the past, say $p$, $p \notin \overline{B}(q,r)$,  MMS is called to find  approximately the shortest path joining $p$ and $q$ along a sequence of bundles of line segments.
\begin{algorithm}[ht]
\caption{{Path planning algorithm for autonomous robot's moving}}
\label{alg:path_planning}
\begin{algorithmic}[1]
\Require A starting point $s$ and a goal $g$ in a map 
\Ensure A path $P$ from $s$ to $g$ in the map.
\State Current point $a \leftarrow s$; next point $a^{next}\leftarrow \emptyset$; initialize the graph $G \leftarrow \emptyset$ to store nodes which are the centers and leaves being endpoints of closed and open sights; the set of open points $OP \leftarrow \emptyset$; the path $P \leftarrow \{s\}$.
\While{True}
\If{$a = g$}		
\State \Return $P$
\Else
\State Get closed and open sights and open points at $a$
\State Add closed sight into $G$
\For{\textbf{each} open point $op$ in open points at $a$}
\If{$op$ is not inside any closed or open sight in $G$}
\State Ranking $op$ by~(\ref{eq:rank})
\State Add $op$ into $OP$
\State Add open sight corresponding to $op$ into $G$
\EndIf
\EndFor
\State $a^{next} \leftarrow$ an open point which is the highest priority rank in $OP$
\If{$a^{next} = \emptyset$}
\State \Return $\emptyset$
\Comment{{\color{blue}there is no path from $s$ to $g$}}
\Else 
\State Find the  shortest path joining $a$ and $a^{next}$ in the graph $G$, denoted by $\tau$.
\State Construct a sequence $\mathcal{F}$ of bundles of line segments attaching with $\tau$ in which two distinct bundles do not intersect each other, the angle of the sector containing each bundle excepting $a$ and $a^{next}$ does not exceed $\pi$\;
\State Call Alg.~\ref{alg:main} for finding approximately the shortest path joining $a$ and $a^{next}$ along $\mathcal{F}$\;
\State $a \leftarrow a^{next}$
\State Remove $a^{next}$ from $OP$
\State Add endpoints of the obtained path into $P$
\EndIf
\EndIf
\EndWhile
\State \Return \textbf{$P$}
\end{algorithmic}
\end{algorithm}
\section{Numerical Experiments}
\label{sect:numerical_results}

\begin{figure}[ht!]
\centering
\begin{subfigure}{.33\textwidth}
\centering
\includegraphics[scale=0.34]{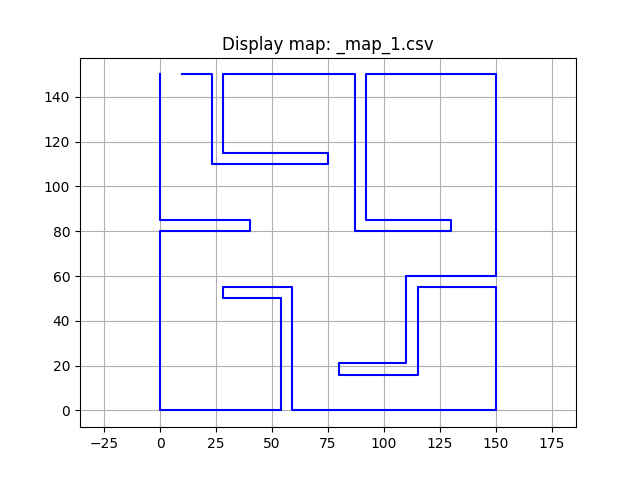}
\caption*{(i)}
\end{subfigure}%
\begin{subfigure}{.33\textwidth}
\centering
\includegraphics[scale=0.34]{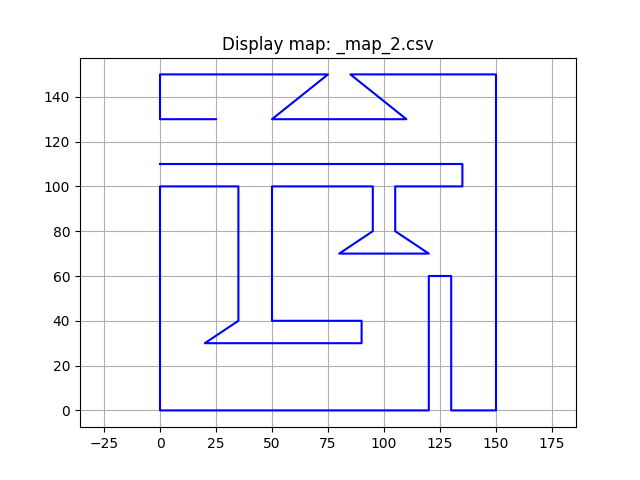}
\caption*{(ii)}
\end{subfigure}
\begin{subfigure}{.3\textwidth}
\centering
\includegraphics[scale=0.34]{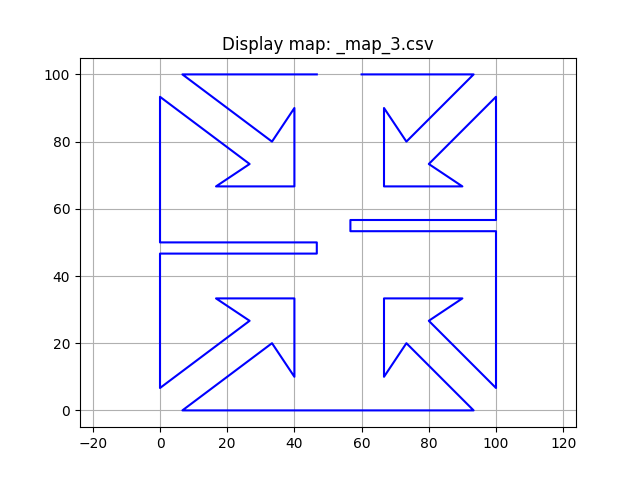}
\caption*{(iii)}
\end{subfigure}
\begin{subfigure}{.33\textwidth}
\centering
\includegraphics[scale=0.34]{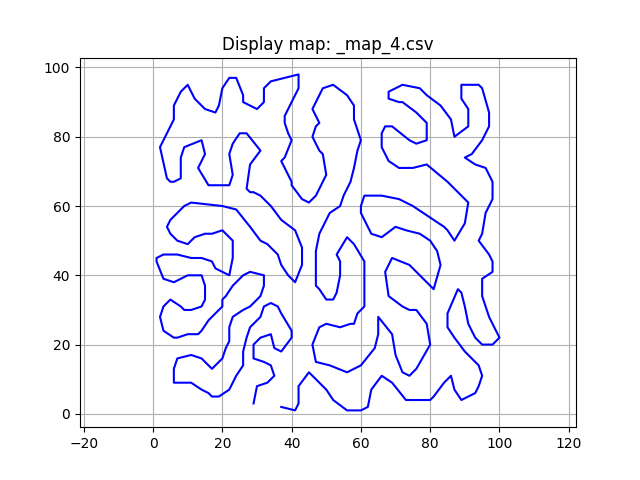}
\caption*{(iv)}
\end{subfigure}%
\begin{subfigure}{.33\textwidth}
\centering
\includegraphics[scale=0.34]{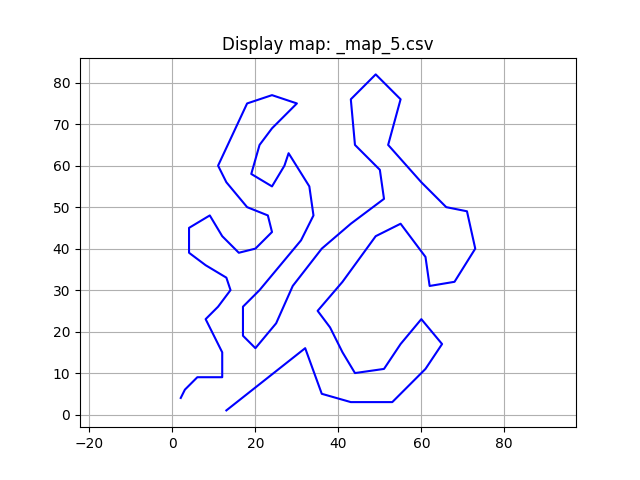}
\caption*{(v)}
\end{subfigure}
\begin{subfigure}{.3\textwidth}
\centering
\includegraphics[scale=0.34]{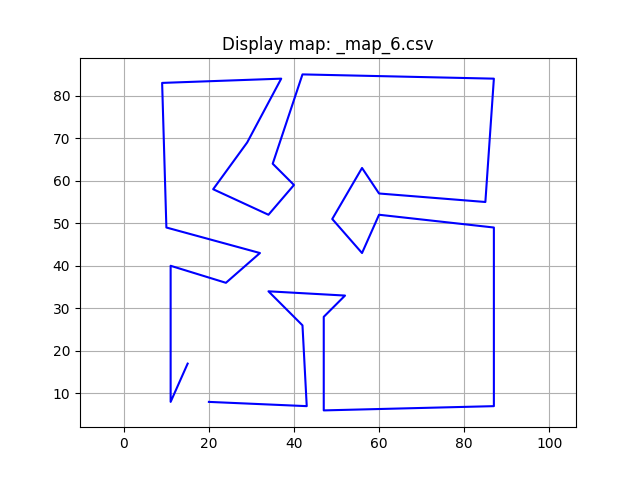}
\caption*{(vi)}
\end{subfigure}
\caption{Some types of map for algorithms mentioned, where distinguishing between walls and other obstacles is unnecessary since, either way, the robot will go around them. These maps are randomly generated by given points on the plane. The respective map sizes are 45, 34, 20, 340, 75, and 30 points.}
\label{fig:maps}
\end{figure}

\begin{figure}[ht]
\centering
\begin{subfigure}{.33\textwidth}
\centering
\includegraphics[scale=0.34]{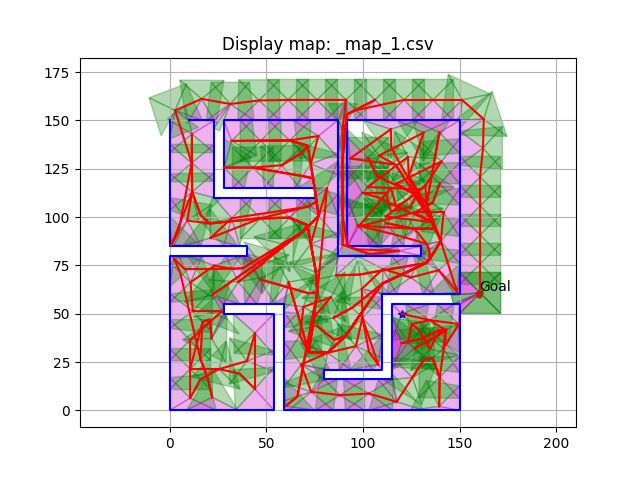}
\caption*{(i): Starting point $(120,50)$ \\ and goal $(160,60)$ with $r=15$}
\end{subfigure}%
\begin{subfigure}{.33\textwidth}
\centering
\includegraphics[scale=0.34]{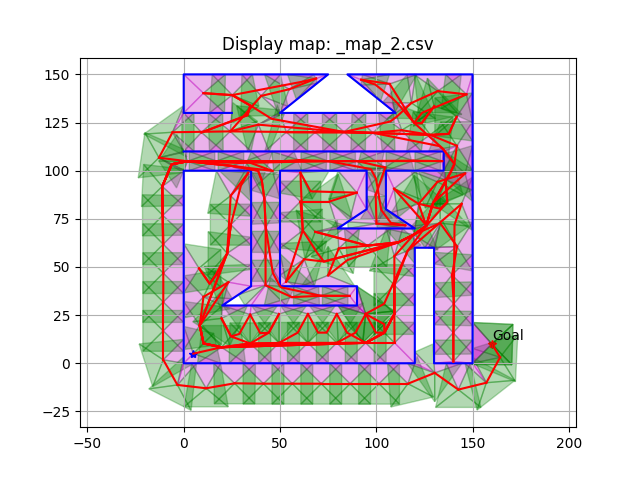}
\caption*{(ii): Starting point $(5,5)$ \\ and goal $(160,10)$ with $r=15$}
\end{subfigure}
\begin{subfigure}{.3\textwidth}
\centering
\includegraphics[scale=0.34]{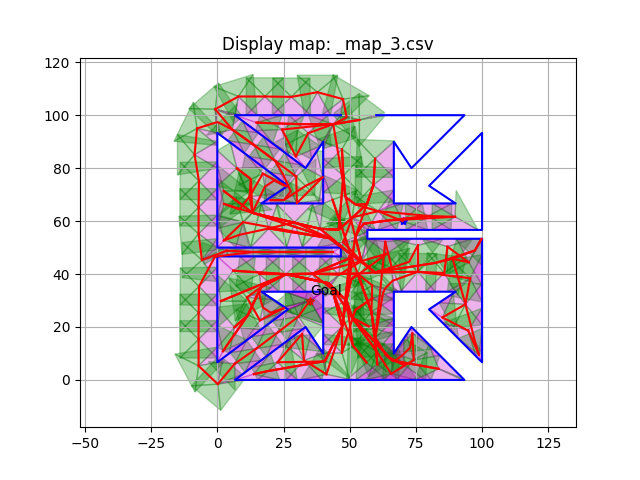}
\caption*{(iii): Starting point $(75,60)$ \\ and goal $(35,30)$ with $r=10$}
\end{subfigure}
%
\begin{subfigure}{.33\textwidth}
\centering
\includegraphics[scale=0.34]{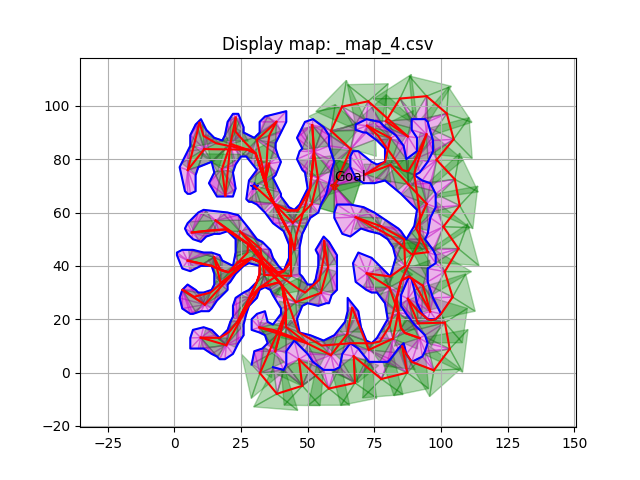}
\caption*{(iv): Starting point $(30,70)$ \\ and goal $(60,70)$ with $r=10$}
\end{subfigure}%
\begin{subfigure}{.33\textwidth}
\centering
\includegraphics[scale=0.34]{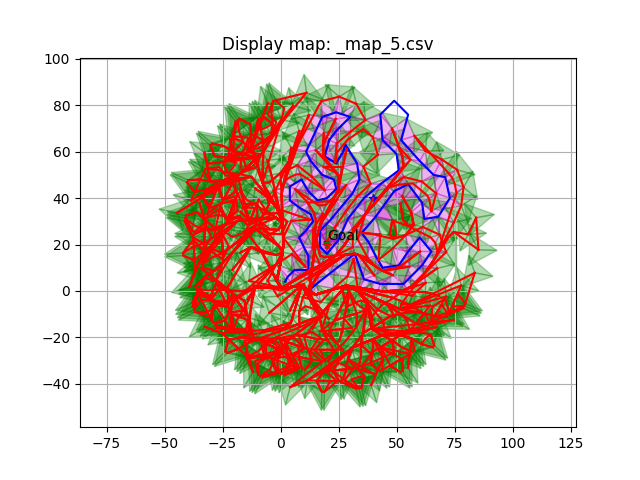}
\caption*{(v): Starting point $(40,40)$ \\ and goal $(20,20)$ with $r=8$}
\end{subfigure}
\begin{subfigure}{.3\textwidth}
\centering
\includegraphics[scale=0.34]{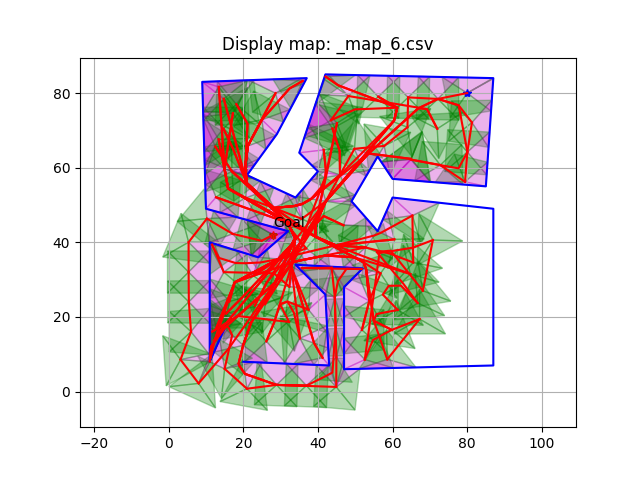}
\caption*{(vi): Starting point $(80,80)$ \\ and goal $(28,42)$ with $r=8$}
\end{subfigure}
\caption{Some motions of robots on some maps obtained by Algorithm~\ref{alg:path_planning} using MMS}
\label{fig:result_maps_10}
\end{figure}

We know that the  rubber band technique of Li and Klette~\cite{Li2011} is  used for finding approximately shortest path along pairwise disjoint line segments whereas line segments of  Problem (P) may be having common points. In implementation, however, An et al.~\cite{BinhAnHoai2021} trimmed line segments of each bundle from the vertex a sub segment of length $\epsilon$ to guarantee that all line segments of bundles are pairwise disjoint,
%
where $\epsilon$ is a very small positive number. 
Then Li and Klette's technique still works.
%
%
%
We test on sets of different maps  as shown in Figure~\ref{fig:maps} and   starting and ending points given by Figure~\ref{fig:result_maps_10}.  For local path planning, we  do not need to differentiate between walls and other obstacles since, either way, the robot will go around them.
The algorithms are implemented in Python and run on  Ubuntu Linux platform Intel Core i5-7200U, CPU 568 2.50GHz.
%
The robot's vision radii are taken respectively as $r=8, r=10$, and $r=15$. 
By the path planning algorithm in Section~\ref{sect:refer_problem} (Algorithm~\ref{alg:path_planning}),
when the robot moves from $q$ to $p$, $p \notin \overline{B}(q,r)$, we obtain a sequence of more than two bundles of line segments attaching with the shortest path joining $p$ and $q$ in the graph $G$, where the sectors containing these bundles excepting $p$ and $q$  do not exceed $\pi$. Let  the number of bundles of line segments of the sequence be $N+2$ ($N >0$).
For each degenerate bundle $a_i (1 \le i \le N)$ such that the sector of $\overline{B}(a_i,r)$ corresponding to the triangle $a_{i-1} a_{i} a_{i+1}$ determines one open sight, for convenience,  we insert the so-called fake segment whose endpoints are $a_i$ and the midpoint of the arc w.r.t the mini sector of $\overline{B}(a_i,r)$. 
Preprocessing gives a sequence of bundles of line segments in which these bundles do not intersect each other.
MMS is then called to find  approximately the shortest path joining two points along the sequence of bundles of line segments.

 To be easy to implement, we partition uniformly such that the number of bundles of each sub-sequence $\mathcal{F}_i (i=0,1,\ldots,K-1)$ are the same and equal to a constant $c$ ($c \in \mathbb{N}, 1 \le c \le N$). The number of bundles of the last sub-sequence $\mathcal{F}_{K}$ can be less than or equal to $c$.  And thus $K$ is the floor of $N/c$ (i.e., $K=\lfloor N/c \rfloor$). The experiments reported in the section are executed with $c = 5$ (i.e., $K=\lfloor N/5 \rfloor$). We also experimented with $c = 6$ (i.e., $K=\lfloor N/6 \rfloor$), but the overall results are generally not significantly different.
%

 To check whether Collinear Condition (A1-A2) holds or not at a shooting point $s_i$,
we use one of two following ways:  to calculate
	the angle created by $\text{SP}( s_{i-1}, s_i)$ and $\text{SP}(s_i, s_{i+1})$ or to determine if $s_i^{next}$ coincides with $s_i$ or not.  By Proposition~\ref{prop:two-stop-condition}, these ways are equivalent. Here we use the latter.
For this determination, we introduce a tolerance denoted as $\epsilon$ to check for the coincidence of points during implementation. Collinear Condition (A1-A2) holds at $s_i$  if $ \Vert s_i^{next} - s_i \Vert < \epsilon$, for all $1 \le i \le K$.

\begin{figure}[ht]
\centering
\begin{subfigure}{.33\textwidth}
\centering
\includegraphics[scale=0.34]{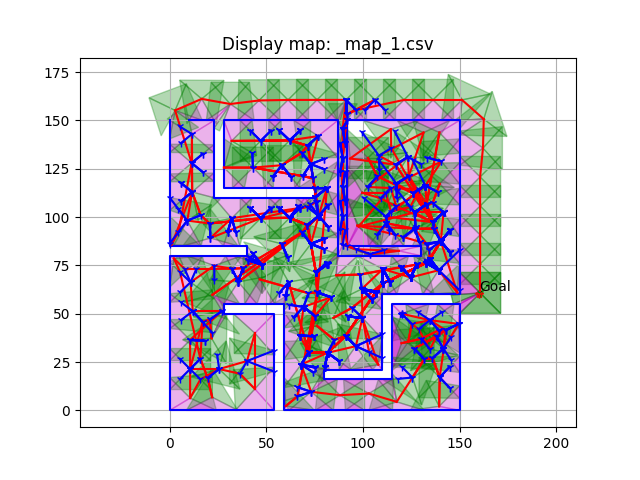}
\caption*{(i): Starting point $(120,50)$ \\ and goal $(160,60)$ with $r=15$}
\end{subfigure}%
\begin{subfigure}{.33\textwidth}
\centering
\includegraphics[scale=0.34]{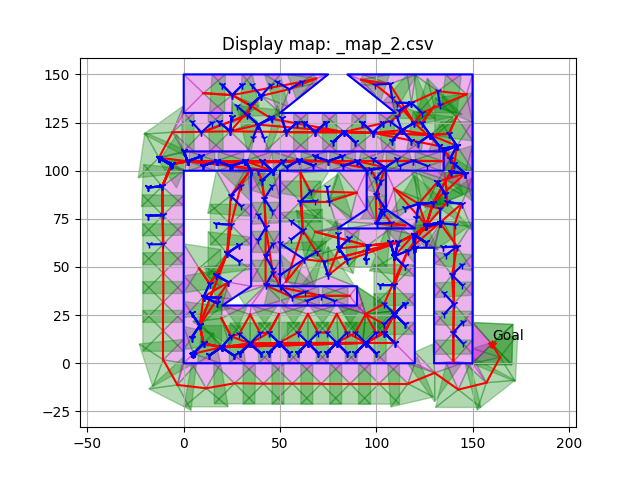}
\caption*{(ii): Starting point $(5,5)$ \\ and goal $(160,10)$ with $r=15$}
\end{subfigure}
\begin{subfigure}{.3\textwidth}
\centering
\includegraphics[scale=0.34]{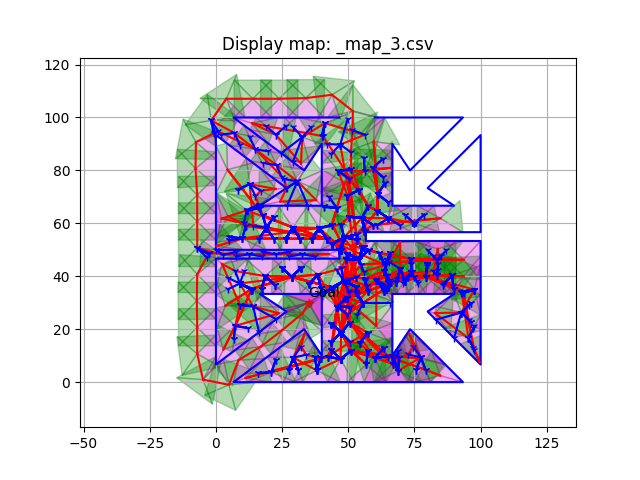}
\caption*{(iii): Starting point $(75,60)$ \\ and goal $(35,30)$ with $r=10$}
\end{subfigure}
%
\begin{subfigure}{.33\textwidth}
\centering
\includegraphics[scale=0.34]{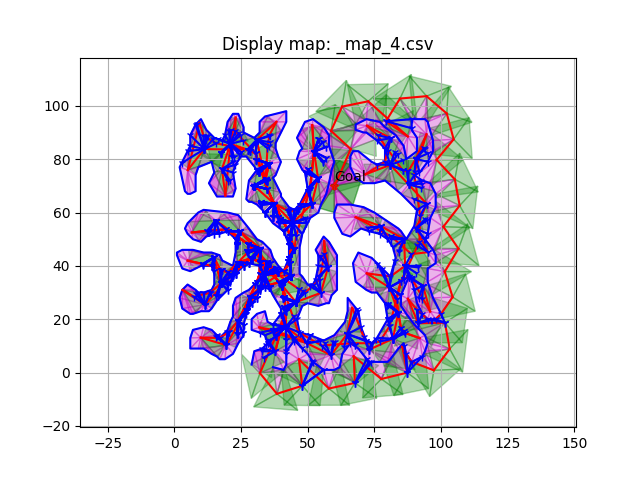}
\caption*{(iv): Starting point $(30,70)$ \\ and goal $(60,70)$ with $r=10$}
\end{subfigure}%
\begin{subfigure}{.33\textwidth}
\centering
\includegraphics[scale=0.34]{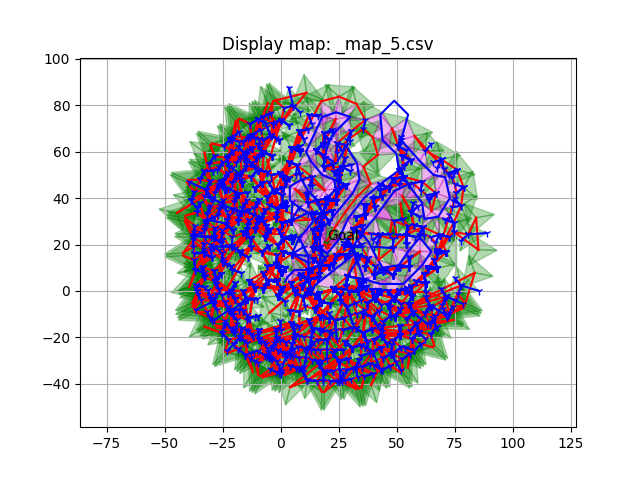}
\caption*{(v): Starting point $(40,40)$ \\ and goal $(20,20)$ with $r=8$}
\end{subfigure}
\begin{subfigure}{.3\textwidth}
\centering
\includegraphics[scale=0.34]{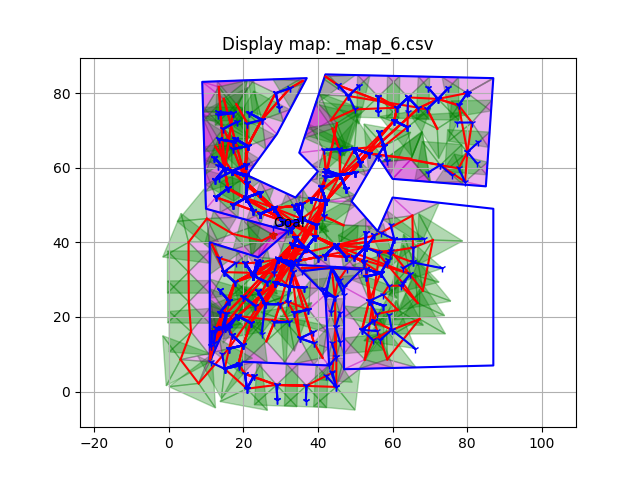}
\caption*{(vi): Starting point $(80,80)$ \\ and goal $(28,42)$ with $r=8$}
\end{subfigure}
\caption{Some sequences of bundles of line segments  created by Algorithm~\ref{alg:path_planning} using MMS}
\label{fig:map_sequence_bundles}
\end{figure}

For comparisons, our implementation is based partly on the source code for path planning of autonomous robots written by An et al.~\cite{BinhAnHoai2021}. 
%
%
Figure~\ref{fig:result_maps_10} demonstrates some motions of an autonomous robot obtained by Algorithm~\ref{alg:path_planning}. Sequences of bundles of line segments obtained by using MMS are shown in Figure~\ref{fig:map_sequence_bundles}.
Tables~\ref{Table:time-results}  describes the  running times (in seconds) of the path planning algorithm  using Li and Klette's technique, Algorithm~\ref{alg:path_planning}  using MMS, and the algorithm using available trajectories obtained from the graph $G$.
%
%

The numerical experiment as in Table~\ref{Table:time-results} 
shows that Algorithm~\ref{alg:path_planning} using MMS averagely reduces the running time by about 2.88\% of the path planning algorithm using Li and Klette's technique, although both approaches get  equivalent lengths of paths traveled by the robot. %
 If we consider the overall context of the autonomous robot problem, which includes  ranking active points to determine the prioritized direction, finding available trajectories from the graph, and calculating the shortest geometric path along a sequence of bundles of line segments, the reduction is negligible. However, when focusing geometric shortest path problem, our algorithm outperforms the other significantly.
In particular, if only considering solving the SP problem for bundles of line segments, the running time of Algorithm~\ref{alg:path_planning} using MMS significantly decreases by  about 64,46\% of that using the technique of Li and Klette. The average speedup ratio of MMS is about 5.49. 

Besides,	Tables~\ref{Table:length-results} 
reports on the length of the robot's paths in the comparison between two approaches: the path planning algorithm using  shortest paths in  $G$  and algorithms replacing these paths with approximate solutions of Problem (P) (the algorithm using Li and Klette's technique, Algorithm~\ref{alg:path_planning} using MMS) for  returning. 
The results  show that the total distance traveled by the robot averagely reduces by  about 16.57\% when applying MMS for finding approximately  shortest paths along sequences of line segments which are used instead of  shortest paths in  graphs. Therefore going back by  shorter paths than  the available trajectories obtained from the graph $G$ save processing time for actual robots.

\begin{figure}[ht]
	\centering
	\begin{subfigure}{.48\textwidth}
		\centering
		\includegraphics[scale=0.64]{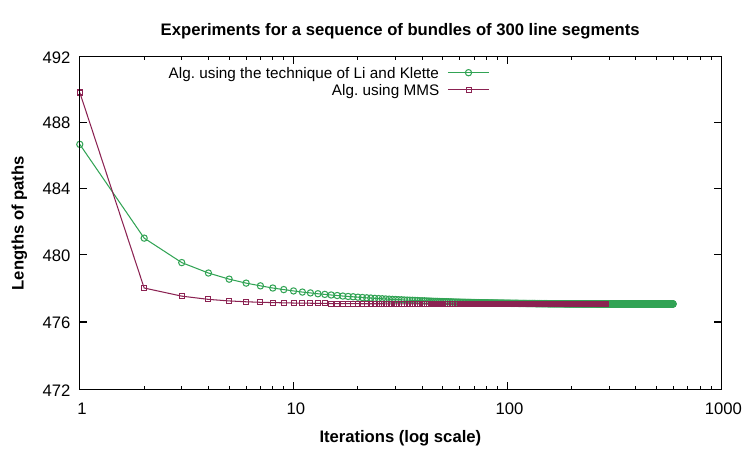}
	\end{subfigure}%
	\begin{subfigure}{.48\textwidth}
		\centering
		\includegraphics[scale=0.64]{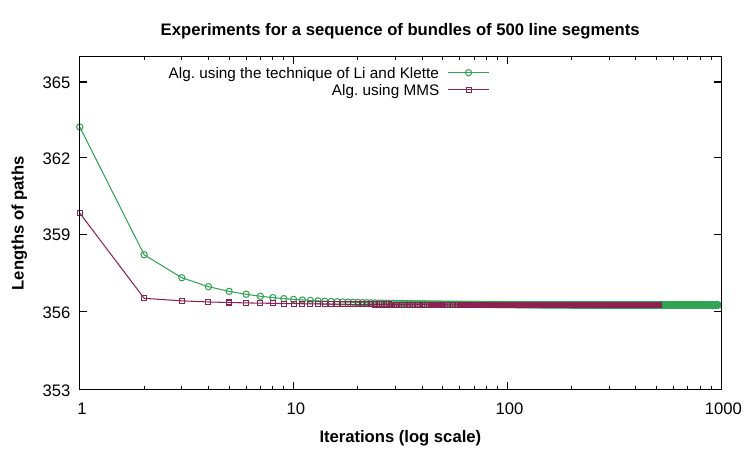}
	\end{subfigure}%
	
	\begin{subfigure}{.48\textwidth}
		\centering
		\includegraphics[scale=0.64]{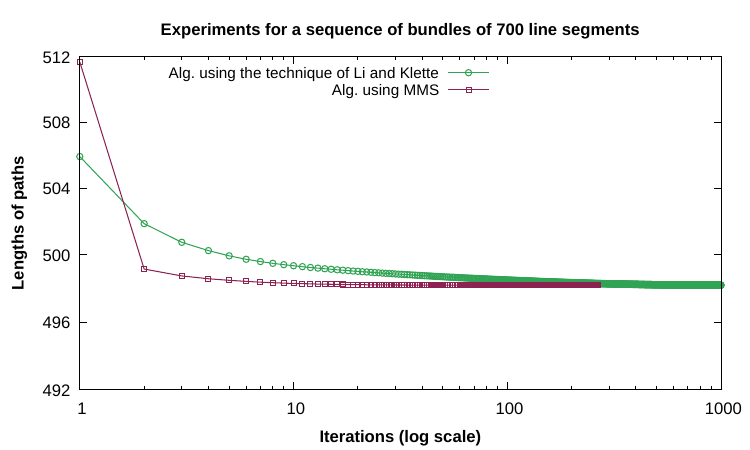}
	\end{subfigure}
	\begin{subfigure}{.48\textwidth}
		\centering
		\includegraphics[scale=0.64]{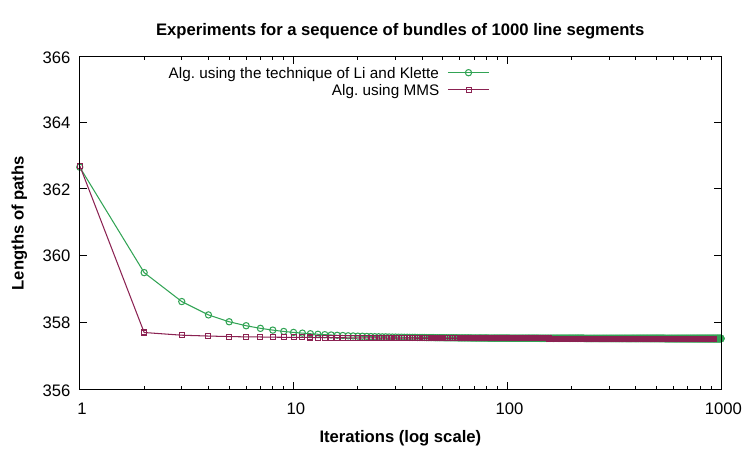}
	\end{subfigure}
	\caption{Comparisons of the lengths of paths obtained after iterations between MMS and the technique of Li and Klette for some data sets.}
	\label{fig:length_compare}
\end{figure}

\begin{figure}[ht]
\centering
\begin{subfigure}{.48\textwidth}
\centering
\includegraphics[scale=0.64]{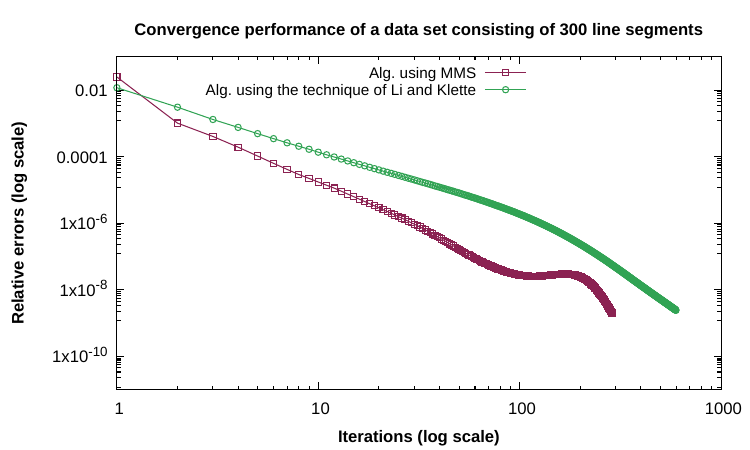}
\end{subfigure}%
\begin{subfigure}{.48\textwidth}
\centering
\includegraphics[scale=0.64]{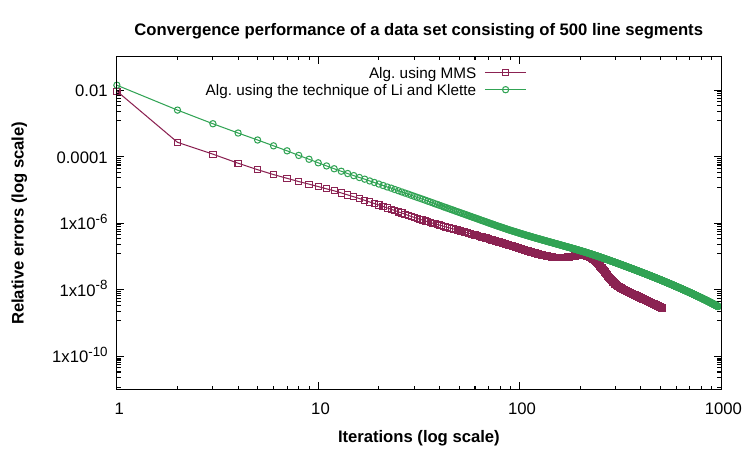}
\end{subfigure}%

\begin{subfigure}{.48\textwidth}
\centering
\includegraphics[scale=0.64]{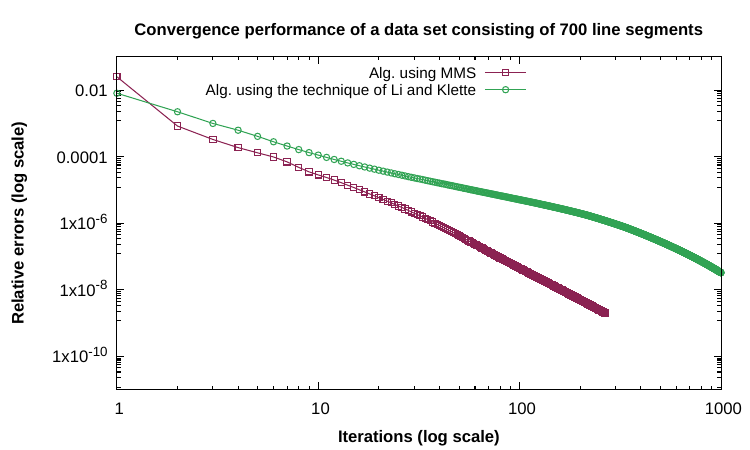}
\end{subfigure}
\begin{subfigure}{.48\textwidth}
\centering
\includegraphics[scale=0.64]{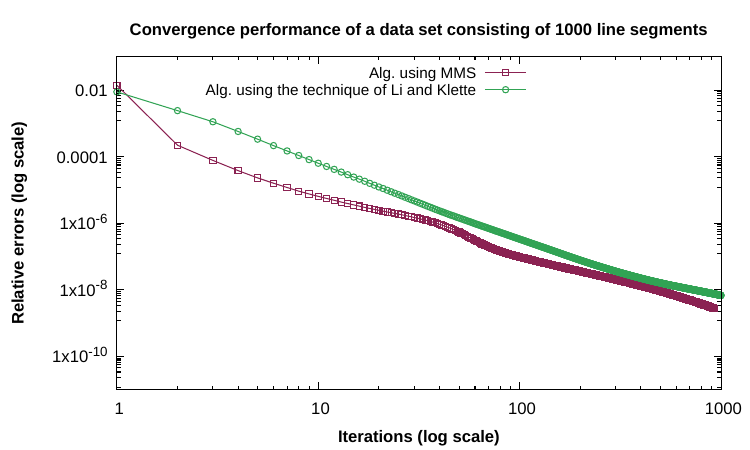}
\end{subfigure}
\caption{Comparisons of the convergence rate between MMS and the technique of Li and Klette for some data sets.}
\label{fig:errors}
\end{figure}

 To compare the convergence of MMS and the technique of Li and Klette, we use relative errors that are defined by: $\dfrac{|l(\gamma_{n})-l(\gamma_{n+1})|}{|l(\gamma_{n+1})|}$, where $l(\gamma_{n})$ is the length of the path obtained after $n^{th}$-step. The iteration of the algorithm using the technique of Li and Klette stops when the relative error at a certain step becomes sufficiently small, whereas MMS stops if the collinear condition holds at all shooting points.
	Figures~\ref{fig:length_compare} and~\ref{fig:errors} show that the convergence rate of the algorithm using MMS is better than that using the technique of Li and Klette for data sets which are sequences of bundles of $300, 500, 700,$ and $1000$ line segments. MMS requires fewer iterations than the technique of Li and Klette to achieve the same relative errors.
	In summary, when solving the SP problem for bundles of line segments, MMS demonstrates faster execution and a more favorable convergence rate compared to Li and Klette's technique.

\section{Conclusion}
\label{sect:conclusion}
In this paper, we have presented the use of MMS with three factors (f1)-(f3) for finding approximately the shortest path joining two points along a sequence of bundles of line segments. These factors involve the partitioning of the sequence of bundles, establishment of the collinear condition, and updating of shooting points.
%
Numerical experiments in Python showed that the 
path planning for autonomous robots in unknown environments using MMS improves running time in the comparison with that using the technique introduced by Li and Klette in~\cite{BinhAnHoai2021}.
%
If we consider the overall context of the autonomous robot problem, 
the reduction of running time is about 2.88\%. However, when focusing on solving the SP problem for bundles of line segments, the running time of Algorithm~\ref{alg:path_planning} using MMS significantly decreases by  about 64,46\% of that using the technique of Li and Klette. Furthermore, the convergence rate of MMS is more favorable  compared to Li and Klette's technique. %
 In our future work, we will concentrate on improving the function of ranking open points to make an autonomous robot move effectively. 
Similarly, MMS can be used efficiently for  autonomous robots moving on terrains, as it works well for bundles in 3D, that topic we will explore in the next paper.


\section*{Acknowledgments}

The authors  thank  Mr. Tran Thanh Binh  (Faculty of Computer Science and Engineering, Ho Chi Minh City University of Technology) for his help in coding Algorithm~\ref{alg:path_planning}.
\par
The  first author would like to thank Ho Chi Minh City University of Technology (HCMUT), VNU-HCM for the support of time and facilities for this study.  We thank
the  anonymous reviewers for their valuable comments on our
manuscrip.
\par

\begin{table}[ht]
\fontsize{10pt}{10pt}
\centering
\caption{ The running times of algorithms: the path planning algorithm using Li and Klette's technique,  Algorithm~\ref{alg:path_planning} using MMS, and the algorithm using available trajectories obtained
from the graph $G$}
\label{Table:time-results}
{\small 
\begin{tabular}{| *{10}{r|}}
	\hline 
	\multicolumn{4}{|c|}{} &   \multicolumn{2}{r|}{} &  Running & \multicolumn{2}{r}{Running time of} &
	\\
	\multicolumn{4}{|c|}{Input}  & \multicolumn{2}{c|}{Running time of} &  time of the & \multicolumn{2}{r}{solving sub-} &
	
	\\
	\multicolumn{4}{|c|}{} &  \multicolumn{2}{r|}{}  &   algorithm & \multicolumn{2}{r}{problems as (P) by} & \\
	
	Map &		Starting  & Goal & Vision & Alg. using &   Alg.~\ref{alg:path_planning} &  using&   Alg. using &   Alg.~\ref{alg:path_planning} & Speedup  \\ 
	&
	point&  & radius&   Li \&  &   using  & available  &   Li \& &  using  &ratio of  \\
	&
	&  &  & Klette's   &   MMS &  trajectories &  Klette's       & MMS  & MMS\\
	
	& & & &technique & & &technique & & \\
	\hline

	\_map\_1 & (120,50) & (160,160) &8&186.3656&182.0621&178.6172&4.9058&1.6197 & 3.03 \\
	\_map\_1 &(5,5)&(140,50)&10&14.2539&13.9766&13.8138&0.1175&0.0599  & 1.96\\
	\_map\_1 &(120,50)&(160,60)&15&39.42&39.2046&38.2075&0.7815&0.3748 &  2.09\\
	\hline
	\_map\_2 &(100,25)&(80,145)&8&166.1297&165.8722&157.0573&4.715&3.4893 & 1.35\\
	\_map\_2 &(10,50)&(30,160)&10&46.0518&45.8075&45.0867&0.39&0.3813 & 1.02\\
	\_map\_2 &(5,5)&(160,10)&15&37.2069&36.8976&36.2428&0.5592&0.3016 & 1.84\\
	\hline
	\_map\_3 &(20,60)&(70,30)&8&54.7647&53.8304&50.2951&1.8857&0.6784 & 2.78\\
	\_map\_3 &(75,60)&(35,30)&10&40.9327&39.4424&37.3475&2.233&0.3932 & 5.68\\
	\_map\_3 &(50,20)&(50,-10)&15&16.1423&15.6189&15.0918&0.6728&0.0578 & 11.64\\
	\hline
	\_map\_4 & (90,70) & (0,60) & 8 & 44.5927 & 44.1078 & 42.6152 & 0.6975 & 0.2499 & 2.79\\
	\_map\_4 &(30,70)&(60,70)&10&56.4147&53.881&51.9044&3.2104&0.2339 & 13.73\\
	\_map\_4 &(95,60)&(40,50)&15&31.3389&27.8889&26.1923&3.7107&0.188 & 19.74\\
	\hline
	\_map\_5 &(40,40)&(20,20)&8&225.1577&224.7189&215.2458&3.7956&1.9902 & 1.91\\
	\_map\_5 &(60,50)&(20,45)&10&9.5332&8.9903&8.3139&0.6873&0.1059 & 6.49\\
	\_map\_5 &(20,55)&(30,40)&15&13.4351&13.1511&12.7288&0.3272&0.0446 & 7.33\\
	\hline
	\_map\_6 &(80,80)&(28,42)&8&31.8206&30.1569&27.7515&0.9105&0.5241 & 1.73\\
	\_map\_6 &(20,50)&(30,60)&10&22.0036&20.6986&19.7666&1.5633&0.1715 & 9.12\\
	\_map\_6 &(20,50)&(50,50)&15&12.8503&12.6942&12.2847&0.2337&0.0506 & 4.62\\
	\hline

\end{tabular} 
}
\end{table}

\begin{table}[ht]
\fontsize{10pt}{10pt}
\centering
\caption{Lengths of paths obtained by the path planning algorithm using available trajectories and
algorithms with replacing available trajectories with shorter paths for returning (the  algorithm using Li and Klette's technique and Algorithm~\ref{alg:path_planning} using MMS)}
\label{Table:length-results}
{\small 
\begin{tabular}{|r|r|r|r|r|r|r|}
	\hline 
	\multicolumn{4}{|c|}{Input} & Lengths of paths  & Lengths of    & Lengths of paths   \\
	Map &		Starting  & Goal & Robot's &  obtained by  & paths obtained   &  obtained  by the \\ &
	point& & vision &   Alg.  using   & by Alg.~\ref{alg:path_planning} &   algorithm using   \\
	&
	& & radius&    Li \& Klette's  &   using MMS &  available  \\
	&
	& & &    technique &    &   trajectories \\
	\hline
	
	\_map\_1 & (120,50) & (160,160) &8&11509.6869&11507.6238&14088.6523  \\
	\_map\_1 &(5,5)&(140,50) &10&1365.673&1365.6031&1647.8448  \\
	\_map\_1 &(120,50)&(160,60)&15&6024.8101&6024.4755&7590.9705 \\
	\hline
	\_map\_2 &(100,25)&(80,145) &8&14063.9094&14063.0613&16868.6556 \\
	\_map\_2 &(10,50)&(30,160)&10&3908.5088&3908.3194&4692.9537 \\
	\_map\_2 &(5,5)&(160,10)&15&4912.8289&4912.5929&5693.1111 \\
	\hline
	\_map\_3 &(20,60)&(70,30)&8&4838.9133&4838.6719&5998.563 \\
	\_map\_3 &(75,60)&(35,30)&10&4353.7386&4353.513&5284.6136 \\
	\_map\_3 &(50,20)&(50,-10)&15&2113.2647&2113.1411&2371.6619 \\
	\hline
	\_map\_4 &(90,70)&(0,60)&8&1633.3712&1632.9317&1849.1196 \\
	\_map\_4 &(30,70)&(60,70)&10&2343.3567&2341.9326&2725.1261 \\
	\_map\_4 &(95,60)&(40,50)&15&2190.4232&2189.8435&2676.0038 \\
	\hline
	\_map\_5 &(40,40)&(20,20)&8&14376.3518&14375.5194&19104.875 \\
	\_map\_5 &(60,50)&(20,45)&10&1067.6096&1067.3361&1225.8483 \\
	\_map\_5 &(20,55)&(30,40)&15&1381.1164&1381.0975&1565.5296 \\
	\hline
	\_map\_6 &(80,80)&(28,42)&8&3906.2147&3905.6391&4787.4772 \\
	\_map\_6 &(20,50)&(30,60)&10&2513.966&2513.1652&3052.9031 \\
	\_map\_6 &(20,50)&(50,50)&15&1753.6594&1753.6194&2131.6867 \\
	
	\hline
	
\end{tabular} 
}
\end{table}


\appendix
\section*{Appendices}
\renewcommand{\thesubsection}{\Alph{subsection}}

\section{The proofs of some propositions for the correctness  of Algorithm~\ref{alg:main}}
\label{apdx:proof}

\label{app.detail-main-alg}


\medskip
\noindent \textbf{Proof of Proposition~\ref{prop:collinear-condition}:}
\begin{proof}
We know that $\gamma$ is the concatenation of shortest paths joining two consecutive shooting points along corresponding sub-sequences of bundles.  
For $i=1,2,\ldots, K$, let $[w_i, s_i]$
and $[s_i,w_i']$ be line segments of $\text{SP}(s_{i-1},s_i)$ and $\text{SP}(s_i,s_{i+1})$, respectively, that have the common endpoint $s_i$. 
Because  Collinear Condition (A1-A2) holds at  $s_i$, we get $\text{SP}(w_i,w'_i) = [w_i, s_i] \cup [s_i,w_i']$, where $\text{SP}(w_i,w'_i)$ is the shortest path joining $w_i$ and $w'_i$ along $\xi_i$s. Since 	
the assumption of Corollary 4.5 in~\cite{HaiAnHuyen2019} is satisfied at  $s_i$, we obtain $\text{SP}(s_{i-1},s_{i+1}) = \text{SP}(s_{i-1},s_i) \cup \text{SP}(s_i,s_{i+1})$, where $\text{SP}(s_{i-1},s_{i+1})$ is the shortest path joining $s_{i-1}$ and $s_{i+1}$ along $\mathcal{F}_{i-1} \cup \mathcal{F}_{i}$.
Therefore $\gamma$ is the shortest path joining $p$ and $q$ along $\mathcal{F}$.
\end{proof}

\medskip

\noindent \textbf{Proof of Proposition~\ref{prop:two-stop-condition}:}
\begin{proof}
$(\Rightarrow)$ Assuming that Collinear Condition (A1-A2) holds  at $s_i$.
%
Let $[w_i, s_i]$
and $[s_i,w_i']$ be line segments of $\text{SP}(t_{i-1},s_i)$ and $\text{SP}(s_i,t_{i})$, respectively, which share $s_i$. 
Because  Collinear Condition (A1-A2) holds at  $s_i$, we get $SP(w_i,w'_i) = [w_i, s_i] \cup [s_i,w_i']$. Since 	
the sufficient condition of Corollary 4.5 in~\cite{HaiAnHuyen2019} is satisfied at  $s_i$, we obtain $\text{SP}(t_{i-1},t_{i}) = \text{SP}(t_{i-1},s_i) \cup \text{SP}(s_i,t_{i})$. Therefore $\text{SP}(t_{i-1}, t_i) \cap [u_i, v_i] = s_i$

$(\Leftarrow)$ Assume that $\text{SP}(t_{i-1}, t_i) \cap [u_i, v_i] = s_i$. 
Then $s_i \in \text{SP}(t_{i-1}, t_i)$.
%
By Theorem 4.1~\cite{HaiAnHuyen2019}, Collinear Condition (A1-A2) holds  at $s_i$. 
\end{proof}

\medskip
\noindent \textbf{Proof of Proposition~\ref{prop:update_path}:}

\begin{proof}
%
If Collinear Condition (A1-A2) holds at all shooting points, 
the proof is straightforward.
Otherwise,  
updating shooting points is shown in (B). 
We deduce that
\begin{align}
\label{eq:update_path_1}
l(\gamma^{current}) = & \sum_{i=0}^{K}l(\text{SP}(s_i,s_{i+1})) \notag \\ 
= & l(\text{SP}(s_0,t_0)) + \sum_{i=0}^{K} \left( l( \text{SP}(t_{i-1},s_i))+l(\text{SP}(s_i, t_{i})) \right) + l(\text{SP}(t_K, s_{K+1})).
\end{align}
As $\text{SP}(t_{i-1},t_{i}) =\text{SP}(t_{i-1},s_i^{next}) \cup \text{SP}(s_i^{next},t_{i})$ and that is the shortest path joining $t_{i-1}$ and $t_{i}$ along $\mathcal{F}_{i-1} \cup \mathcal{F}_{i}$, we obtain
\begin{align}
\label{eq:update_path_2}
l(\text{SP}(t_{i-1},s_i))+l(\text{SP}(s_i, t_{i})) \ge l(\text{SP}(t_{i-1},s_i^{next}))+l(\text{SP}(s_i^{next}, t_{i})).
\end{align}
Combining~(\ref{eq:update_path_1}) with~(\ref{eq:update_path_2}) and note that $s_0^{next}=p, s_{K+1}^{next}=q$, we get
\begin{align}
\label{eq:update_path_3}
l(\gamma^{current}) > & \,  l(\text{SP}(s_0^{next},t_0)) + \sum_{i=0}^{K} \left( l(\text{SP}(t_{i-1},s_i^{next}))+l(\text{SP}(s_i^{next}, t_{i})) \right) \notag \\
& + l(\text{SP}(t_K, s_{K+1}^{next})) \notag \\
= &  \sum_{i=0}^{K} \left( l(\text{SP}(s_i^{next},t_i)) +  l(\text{SP}(t_i,s_{i+1}^{next})) \right).
\end{align}
As $\text{SP}(s_i^{next},s_{i+1}^{next})$ is the shortest path joining $s_i^{next} $ and $s_{i+1}^{next}$ along $\mathcal{F}_i$, we obtain
\begin{align}
\label{eq:update_path_5}
l(\text{SP}(s_i^{next}, t_i))+ l(\text{SP}( t_i, s_{i+1}^{next})) \ge l (\text{SP}(s_i^{next},s_{i+1}^{next})).
\end{align}
Combining~(\ref{eq:update_path_3}) with~(\ref{eq:update_path_5}) yields:
\begin{align*}
l(\gamma^{current}) \ge \sum_{i=0}^{K} \left( l (\text{SP}(s_i^{next},s_{i+1}^{next})) \right)
=  l \left( \cup_{i=0}^{K} \text{SP}(s_i^{next},s_{i+1}^{next}) \right) = l (\gamma^{next}).
\end{align*}

If Collinear Condition (A1-A2) does not hold at some shooting point $s_i$, then the inequality~(\ref{eq:update_path_2}) is strict. As a result, the final inequality is strict. 
\end{proof}

\medskip
\noindent \textbf{Proof of Theorem~\ref{theo:global solution}:}

\begin{proof}
Assume that the number of line segments of $\mathcal{F}$ except $p$ and $q$ is $J$ ($J \ge 1$). We write $\mathcal{F} =\{ e_m = [x_m,y_m], \, m = 1,2, \ldots, J\}$, in which the order of $x_m$ and $y_m$ is as follows:
\begin{itemize}
\item If $e_m$  has an endpoint lying on the right  of  $\tau$ when we traverse from $p$ to $q$, then $x_m$  is this endpoint and $y_m$ is the remaining one.
\item Otherwise, $x_m$ is the vertex of the bundle containing $e_m$  and $y_m$ is the remaining endpoint.
\end{itemize}
Clearly, if  $e_m = \xi_i = [u_i,v_i]$, then $x_m = v_i, y_m = u_i$, for $i=0, 1,\ldots,K+1$.
Let $\mathcal{P}$ be the polyline connecting $p$,  $\{x_m\}_{m=1}^J$ and $q$. Let $\mathcal{Q}$ be the polyline connecting $p$, $\{y_m\}_{m=1}^J$ and $q$.
Due to the assumption of the theorem,
$\mathcal{P} \cup \mathcal{Q}$ is a simple polygon.


Let $\gamma^j = \cup_{i=0}^K \text{SP}(s_i^j,s_{i+1}^j)$, where shooting points $s_i^j $ belong to cutting segments $\xi_i = [u_i, v_i]$ for $i=1,2, \ldots, K$ and $s_0^j = p, s_{K+1}^j =q$. 
If the algorithm stops after a finite number of iterations,  according to Proposition~\ref{prop:collinear-condition}, the shortest path is obtained. We thus just need to consider the case that $\{\gamma^j\}_{j\in \mathbb{N}}$ is infinite.
%
Therefore it is necessary to show that $\gamma^j$ convergences to $\gamma^*$ as  $j \rightarrow +\infty$ in $\mathcal{P} \cup \mathcal{Q}$.
\begin{figure}[ht!]
\centering
\includegraphics[width=.68\linewidth]{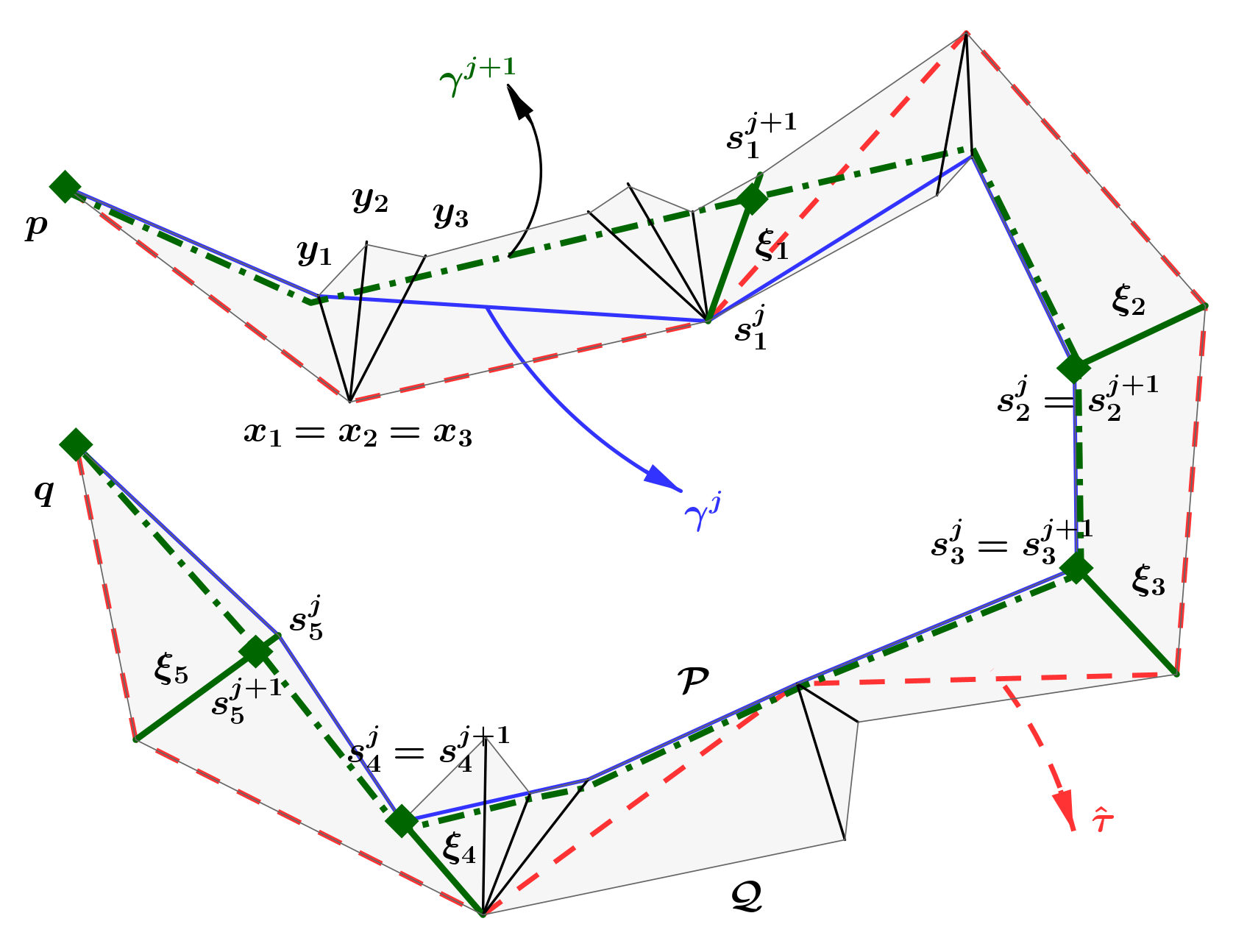}
\caption{Illustration of the proof of Theorem~\ref{theo:global solution}}
\label{fig:proof_theo1}
\end{figure}
The proof is completed by the following claims:
\par
\textbf{Claim 1:} Firstly, by Lemma~\ref{lem:order-gamma-j}, we obtained that the path $\gamma^{j+1}$  at $(j+1)^{th}$-step  are always above the paths $\gamma^j$ at $j^{th}$-step
with respect to the  directed lines  $v_iu_i$, for $i=1,2,\ldots,K$, i.e., $s_i^{j+1} \in [s_i^j,u_i]$, where $s_i^{j+1}$ and $s_i^j$ are shooting points at $(j+1)^{th}$-step and $j^{th}$-step, respectively, 
see Figure~\ref{fig:proof_theo1}.
\par
\textbf{Claim 2:} Next we show that the sequence of shooting points on each cutting segment $[v_i,u_i]$ are convergent to $s_i^*$, for $i =1,2, \ldots K$. Let $\hat{\gamma}$ be the path formed by these limited points $s_i^*$. 
\par
\textbf{Claim 3:} We show that $d_{\mathcal{H}}(\gamma^j,\hat{\gamma}) \rightarrow 0$ and  $l(\gamma^j) \rightarrow l(\hat{\gamma})$ as $j \rightarrow +\infty$.
\par 
\textbf{Claim 4:} Finally, $\hat{\gamma}$ is proved to be the shortest path joining $p$ and $q$ in  $\mathcal{P} \cup \mathcal{Q}$.
\medskip

\textbf{Proof of Claim 1:} \; \textit{See Lemma~\ref{lem:order-gamma-j}.}
Now,
for each $i=1,2,\ldots, K$, 
%
since $[u_i,v_i]$ is compact  and $\{s_i^j\}_{j\in \mathbb{N}} \subset [u_i,v_i]$, there exists a sub-sequence $\{s_i^{j_k}\}_{k \in \mathbb{N}} \subset \{s_i^j\}_{j\in \mathbb{N}}$ such that $\{s_i^{j_k}\}_{k \in \mathbb{N}}$ is convergent to a point, says $s_i^*$, in $\mathbb{R}^2$ with $\Vert \cdot \Vert$ as $k \rightarrow \infty$. 
As $[u_i, v_i]$ is closed and $\{s_i^{j_k}\}_{k \in \mathbb{N}} \subset [u_i,v_i]$, $\{s_i^{j_k}\}_{k \in \mathbb{N}} \rightarrow s_i^*$, we get $s_i^* \in [u_i, v_i]$. 
Write $s_0^*=p$ and $s_{K+1}^*=q$. Set $\hat{\gamma} = \cup_{i=0}^K \text{SP}(s_i^*,s_{i+1}^*)$.

\textbf{Proof of Claim 2:} \; \textit{We begin by proving the convergence of whole sequences, i.e., $\{s_i^j\}_{j\in \mathbb{N}} \rightarrow s_i^*$ as $j \rightarrow \infty$, for $i=1,2, \ldots, K$.}
By the order of elements of $\{s_i^j\}_{j\in \mathbb{N}}$ as shown in Lemma~\ref{lem:order-gamma-j}, for all natural number large enough  $j$, there always exits $k \in \mathbb{N}$ such that $ s_i^j \in [s_i^{j_k},s_i^{j_{k+1}}]$.
Furthermore,  we also obtain $\{s_i^{j_k}\}_{k\in \mathbb{N}}$ converges on one side to $s_i^*$.
For all $\delta >0$, there exists $k_0 \in \mathbb{N}$ such that $\Vert s_i^{j_k} - s_i^* \Vert < \delta$, for $k \ge k_0$.
%
As $s_i^{j} \in [s_i^{j_{k_0}}, s_i^*]$, for $j \ge j_{k_0}$, we get $\Vert s_i^{j} - s_i^* \Vert < \delta$, for $j \ge j_{k_0}$. 
This clearly forces $\{s_i^{j}\}_{j \in \mathbb{N}} \rightarrow s_i^*$ as $j \rightarrow \infty$. This  is the one-sided convergence, $\hat{\gamma}$ is thus above  $\gamma^j$ 
%
with respect to the  directed lines  $v_iu_i$, for $i=1,2,\ldots,K$.

\textbf{Proof of Claim 3:} \; \textit{We next indicate that  $d_{\mathcal{H}}(\gamma^j, \hat{\gamma}) \rightarrow 0$ and  $l(\gamma^j) \rightarrow l(\hat{\gamma})$ as $j \rightarrow \infty$.}
\textit{We indicate that  $d_{\mathcal{H}}(\gamma^j, \hat{\gamma}) \rightarrow 0$ and  $l(\gamma^j) \rightarrow l(\hat{\gamma})$ as $j \rightarrow \infty$.}
Let us denote $\hat{\gamma}_i = {\rm SP} (s_i^*,s_{i+1}^*)$ and $\gamma^j_i = {\rm SP}(s^j_i,s^j_{i+1})$. Then $\hat{\gamma} = \cup_{i=0}^K \hat{\gamma}_i$ and $\gamma^j = \cup_{i=0}^K \gamma^j_i$.
Due to Claim 2,  $s^j_i \rightarrow s_i^*$ as $j \rightarrow \infty$, for $i=1,2,\ldots, K$. 
By Lemma 3.1 in~\cite{HaiAn2011}, 
we get $d_{\mathcal{H}}\left( {\rm SP}(s^j_i,s^j_{i+1}),{\rm SP} (s_i^*,s_{i+1}^*) \right)$  $\rightarrow 0$ as $j \rightarrow \infty$, for $i=1,2,\ldots, K$.  
Note that $d_{\mathcal{H}} (A \cup B, C \cup D) \le \max \{ d_{\mathcal{H}} (A,	C), d_{\mathcal{H}} (B,D)\}$, for all closed sets $A,B,C$ and $D$ in $\mathbb{R}^2$, then
$$d_{\mathcal{H}} (\gamma^j, \hat{\gamma}) \le \max_{0 \le i \le K} \{d_{\mathcal{H}}\left( {\rm SP}(s^j_i,s^j_{i+1}),{\rm SP} (s_i^*,s_{i+1}^*) \right) \}.$$
Therefore $d_{\mathcal{H}}(\gamma^j, \hat{\gamma}) \rightarrow 0$ as $j \rightarrow \infty$.
According to~\cite{HaiAn2011} (p. 542), 
the geodesic distance between two points $x$ and $y$ in a simple polygon, which is measured by the length of the shortest path joining two points $x$ and $y$ in the simple polygon, is a metric on the simple polygon and it is continuous as a function
of both $x$ and $y$. That is, $l({\rm SP}(s_i^j, s_{i+1}^j)) \rightarrow l({\rm SP}(s_i^*, s_{i+1}^*))$ as $j \rightarrow \infty$, for $i=0,1,\ldots, K$. It follows $l(\gamma^j) \rightarrow l(\hat{\gamma})$ as $j \rightarrow \infty$.

%


\textbf{Proof of Claim 4:} \;
By using the  same way of the  proof of Theorem 1  in~\cite{TrangLeAn2021} (item \textbf{iii.}),  we obtain a similar argument that $\hat{\gamma}$ is the shortest path joining $p$ and $q$ in the simple polygon $\mathcal{P} \cup \mathcal{Q}$. Note that the unique difference of the assumption of Theorem~\ref{theo:global solution} in this paper to that in~\cite{TrangLeAn2021} is shootings points maybe coincide with $u_i$ of cutting segment $\xi_i$. 
Summarily, the proof is complete.
\end{proof}

The following lemma  is similar to Proposition 5 in~\cite{TrangLeAn2021}.
\begin{lemma}
\label{lem:order-gamma-j}

For all $j$, we have 
$s_i^{j+1} \in [s_i^j, u_i]$, 
for $i=1,2, \ldots, K$, 
where $s_i^{j+1}$ and $s_i^j$ are shooting points at $(j+1)^{th}$-step and $j^{th}$-step, respectively.

\end{lemma}

\begin{proof}
The proof is done by induction on $j$.
The statement holds for $j=1$ due to taking initial shooting points $s_i^0 =v_i$, for $i=1,2,\ldots, N$.
Let $\{s_i^{prev}\}_{i=1}^N, \{s_i\}_{i=1}^N$ and $\{s_i^{next}\}_{i=1}^N$ be  shooting points corresponding to $(k-1)^{th}$, $k^{th}$ and $(k+1)^{th}$-iteration step, respectively of the algorithm ($k \ge 1$).
Assuming that,  for $i=1,2, \ldots, N$,  $s_i \in [s_i^{prev}, u_i]$, we next prove that $s_i^{next} \in [s_i, u_i]$.

%
If  $\angle_{v_iu_i} \left( \text{SP}(s_{i-1},s_i), \text{SP}(s_i,s_{i+1}) \right) \le \pi$, we  obtain  $s_i^{next} \in [s_i, u_i]$. Thus we just need to prove the case $\angle_{v_iu_i} \left( \text{SP}(s_{i-1},s_i), \text{SP}(s_i,s_{i+1}) \right) > \pi$.
If $s_i = v_i$, then Collinear Condition (B) holds at  $s_i$ and therefore $s_i^{next} = s_i \in [s_i, u_i]$.
If $s_i \neq v_i$,  suppose contrary to our claim,  $\angle_{v_iu_i} \left( \text{SP}(s_{i-1},s_i), \text{SP}(s_i,s_{i+1}) \right) > \pi$ and $s_i^{next} \notin [s_i, u_i]$, i.e., 
\begin{align}
\label{eq:7}
s_i^{next} \in [v_i, s_i[ 
\end{align}

Let $G_1$  be the closed region bounded by $[s_{i}^{prev},s_i]$, $\text{SP}(s_i,t_{i-1})$ and $\text{SP}(t_{i-1},s_{i}^{prev})$. Let $G_2$  be the closed region bounded by $[s_{i}^{prev},s_i]$, $\text{SP}(s_i, t_{i})$ and $\text{SP}(t_{i},s_{i}^{prev})$, where $t_{i-1}$ and $t_{i}$ are  defined
in (B)  (see Fig.~\ref{fig:prooflemma}). Since $s_i \in [u_i,v_i[$, $\angle_{v_iu_i} \left( \text{SP}(t_{i-1},s_i), \text{SP}(s_i,t_{i}) \right) \le \pi$.
%
Next, let $[s_i,w_1]$ and $[s_i, w_2]$ be segments having
the common endpoint $s_i$ of $\text{SP}(s_{i-1},s_i)$ and $\text{SP}(s_i,s_{i+1})$, respectively.
\begin{figure}
\centering
\includegraphics[width=0.6\linewidth]{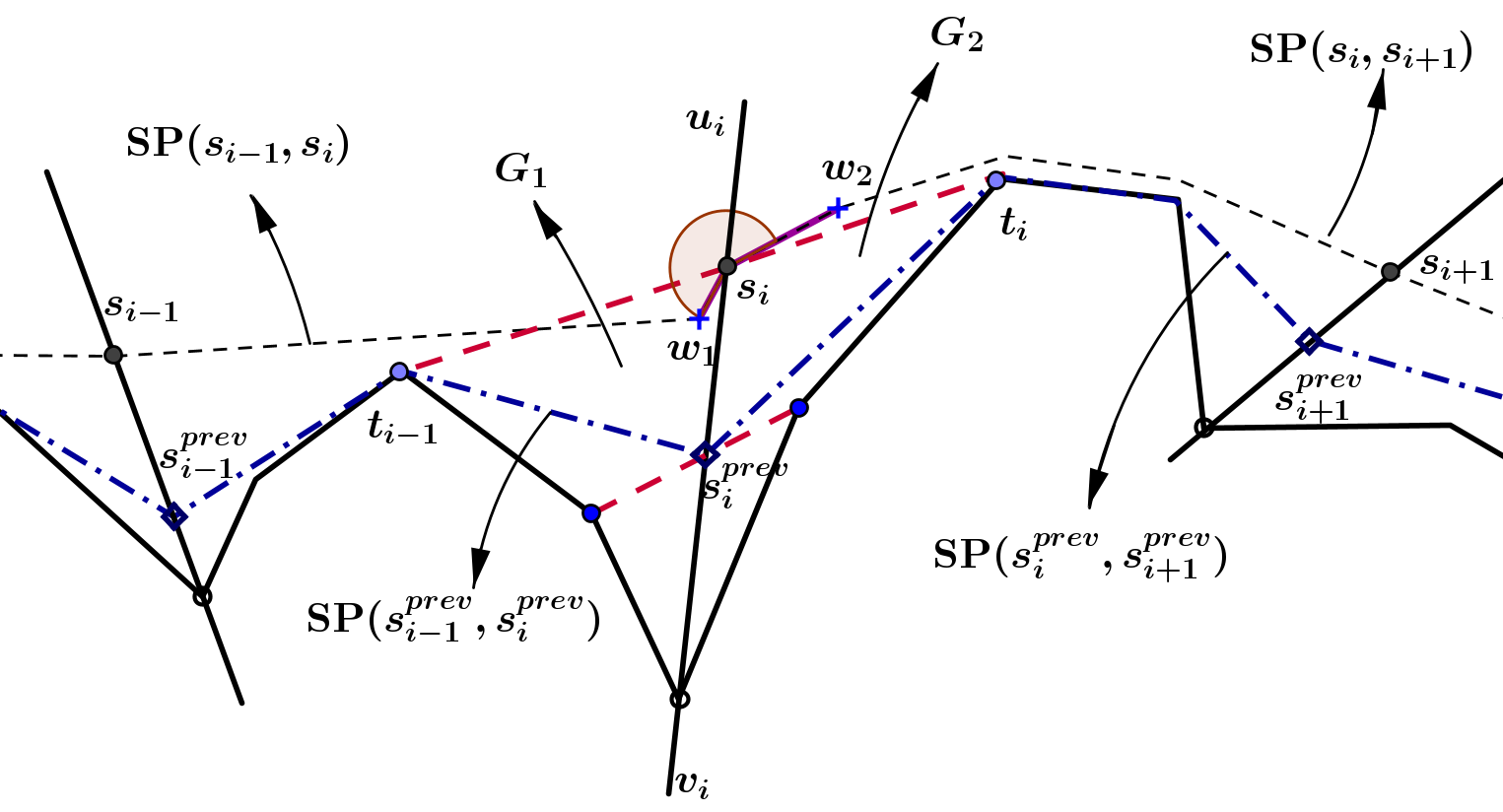}
\caption{Illustration of the proof of Lemma~\ref{lem:order-gamma-j} for the crossing path case at $s_i$}
\label{fig:prooflemma}
\end{figure}
As  $\angle_{v_iu_i} \left( \text{SP}(t_{i-1},s_i), \text{SP}(s_i,t_{i}) \right) \le \pi$ and  $\angle_{v_iu_i} \left( \text{SP}(s_{i-1},s_i), \text{SP}(s_i,s_{i+1}) \right) > \pi$,  at least one of the following cases happens: 
\begin{align}
\label{eq:two-cases-cross}
&	]s_i, w_1]\text{ is below entirely  } \text{SP}(t_{i-1},s_i) \\ \notag
&	\text{    or } ]s_i, w_2] \text{  is below entirely } \text{SP}(s_i, t_{i}) \text{  w.r.t. the  directed line joining } v_i \text{  and } u_i. \hspace{4cm}
\end{align}

Obviously, $\text{SP}(s_i, t_{i})$, $\text{SP}(s_i,s_{i+1})$ and $\text{SP}(s_i^{prev},t_i)$ do not cross each other, due to the properties of shortest paths.
For all $t_i \in \text{SP}(s_i^{prev},s_{i+1}^{prev})$, $\text{SP}(s_i, t_{i})$ is entirely contained in the polygon bounded by $[s_i^{prev}, s_i]$,  $[s_{i+1}, s_{i+1}^{prev}]$ $\text{SP}(s_i,s_{i+1})$ and $\text{SP}(s_i^{prev},s_{i+1}^{prev})$. Thus $\text{SP}(s_i,s_{i+1})$ does not intersect with the interior of $G_2$. Similarly, $\text{SP}(s_{i-1}, s_i)$ does not intersect with the interior of $G_1$. These things contradict~(\ref{eq:two-cases-cross}), which  completes the proof.
\end{proof}

%

\medskip
\noindent \textbf{Proof of Proposition~\ref{theo:complexity}:}
\begin{proof}
	Step~\ref{step1} of Algorithm~\ref{alg:main} needs $O(1)$ time.
Note that the number of bundles of $\mathcal{F}_i (i =0,1,\ldots,K)$ does not exceed $c$.
According to Section~\ref{apdx:example} in Appendices, for a sub-sequence $\mathcal{F}_i (i =0,1,\ldots, K)$, the shortest paths joining two shooting points along $\mathcal{F}_i$ is calculated by using the funnel technique no more than $2^{c-1}$ times. These paths correspond to the shortest paths along no more than $2^{c-1}$ sequences of adjacent triangles. Since the funnel technique gives linear time complexity and $c$ is a constant, it follows that  computing $\gamma^{current}$ and $\gamma^{next}$, which are formed  by sets of shooting points, takes $O(J)$ time.
Hence, steps~\ref{step2}, and~\ref{step3} of Algorithm~\ref{alg:main} need $O(J)$ time.
Now, we calculate the time complexity of Procedure {\sc Collinear\_Update}($\gamma^{current}, \gamma^{next}, flag$).
For each iteration step, 
then computing sequentially the angles at shooting points is done in  constant time. In  steps~\ref{prcRobot:step8},~\ref{prcRobot:step14}, and~\ref{prcRobot:step20} of for-loop in Procedure {\sc Collinear\_Update},  since there are at most two procedures for the SP problem for bundles along $\mathcal{F}_{i-1} \cup \mathcal{F}_i$ which call each line segments of $\mathcal{F}$,
time complexity of Procedure {\sc Collinear\_Update}($\gamma^{current}, \gamma^{next}, flag$) is $O(KJ)$, where $K$ is the number of cutting segments. As $K = \lfloor N/c \rfloor$, the complexity of the procedure is $O(NJ)$, and thus step~\ref{iteration} of Algorithm~\ref{alg:main} also runs in $O(NJ)$ time.
In summary, the algorithm runs in $O(mNJ)$ time, where $m$ is the number of iterations to get the required path.
\hfill
\end{proof}

\begin{figure}[ht!]
\centering
\includegraphics[width=0.5\linewidth]{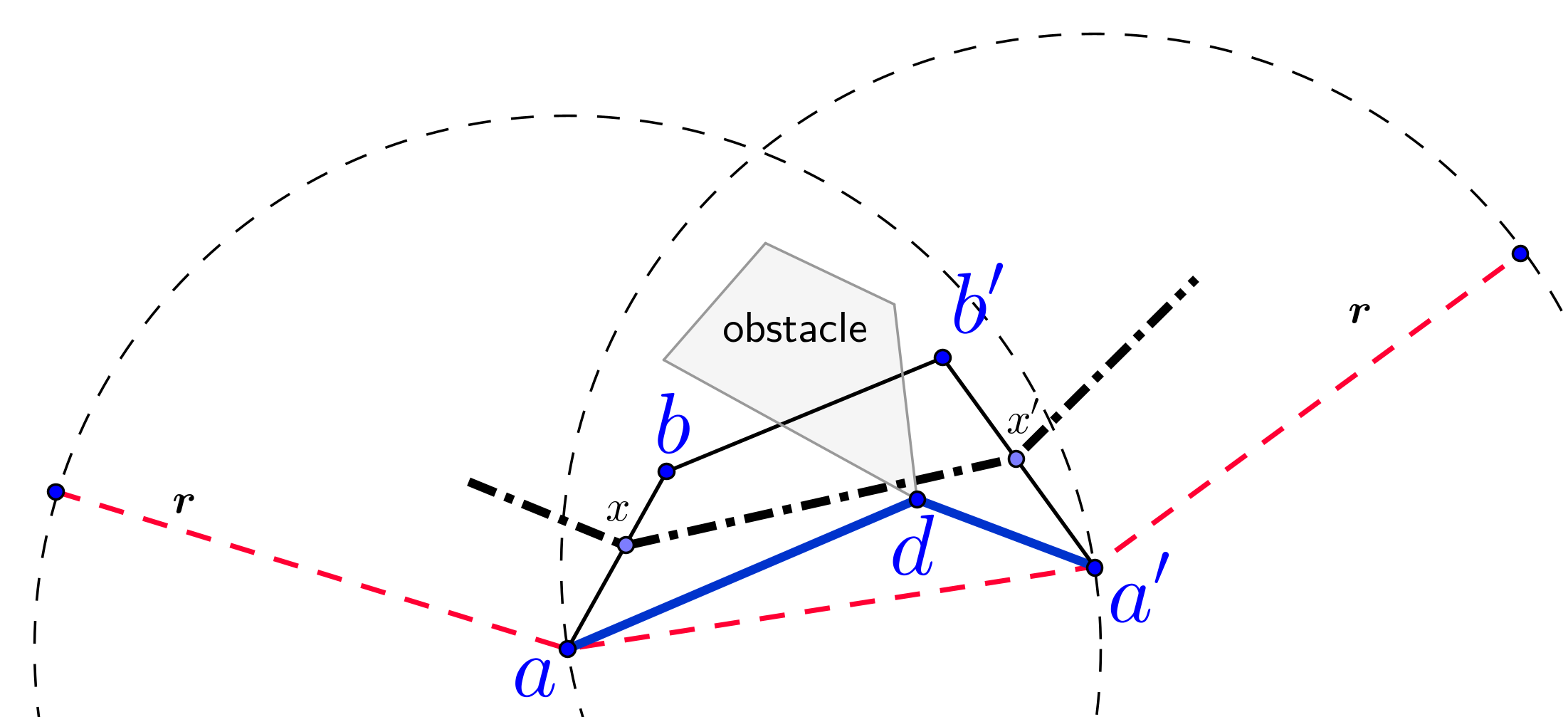}
\caption{Illustration of proof of Proposition~\ref{obser:reduce_cross}}
\label{fig:minimum_polygon}
\end{figure}

\medskip
\noindent \textbf{Proof of Proposition~\ref{obser:reduce_cross}:}
\begin{proof}
(a) 
%
Recall that $\hat{\mathcal{F}} =${\sc Perprocessing}$(\mathcal{F}^*)$, where $\mathcal{F}^* = \{ a_0, a_1, \ldots, a_{N+1}\}$.
When we split bundles of line segments attaching with $a_i$ by the boundary of $\overline{B}(a_i,r_0)$, 	two distinct bundles of  $\hat{\mathcal{F}}$ do not intersect  each other, where $r_0 \le\frac{1}{2} \min\limits_{\substack{ 0 \le j,k \le N \\ j\neq k}} \Vert a_j - a_k\Vert$. Thus (a) holds.

(b) Assume that the shortest path joining two points along  $\hat{\mathcal{F}}$ does not avoid the obstacles.
Then the  path contains at least a segment, say $[x,x']$, such that $[x,x']$ intersects the interior of an obstacle, where $x$ and $x'$ belong to two consecutive line segments of $\hat{\mathcal{F}}$. Clearly  these two  line segments belong to two different bundles. Let us denote these two  line segments by $[a,b]$ and $[a',b']$, where $a$ and $a'$ are two consecutive bundles.
According to preprocessing bundles, $\Vert a-b \Vert \le r_0 \le r/2$ and  $\Vert a'-b' \Vert \le r_0 \le r/2$. Therefore, the quadrangle $abb'a'$ included $\overline{B}(a,r) \cup  \overline{B}(a',r)$, see Figure~\ref{fig:minimum_polygon}. Since $[x,x']$ intersects the interior of an obstacle, there is at least a vertex of this obstacle, say $d$, which is inside $abb'a'$. Due to the construction of closed sights and line segments of $\mathcal{F}^*$, then $[a,d]$ or $[a',d]$ are also line segments of $\mathcal{F}^*$. Thus $[a,d] \cap \overline{B}(a,r_0)$ and $[a',d] \cap \overline{B}(a',r_0)$ are also line segments of $\hat{\mathcal{F}}$ and they are different to $[a,b]$ and $[a',b']$. This contradicts the assumption that $[a,b]$ and $[a',b']$ are two consecutive line segments of $\hat{\mathcal{F}}$.

\end{proof}

\section{Some drawbacks of the previous algorithms for the SP problem for bundles}
\label{apdx:example}
\begin{figure}[htp]
\begin{subfigure}[t]{.48\textwidth}
\centering
\includegraphics[scale=0.35]{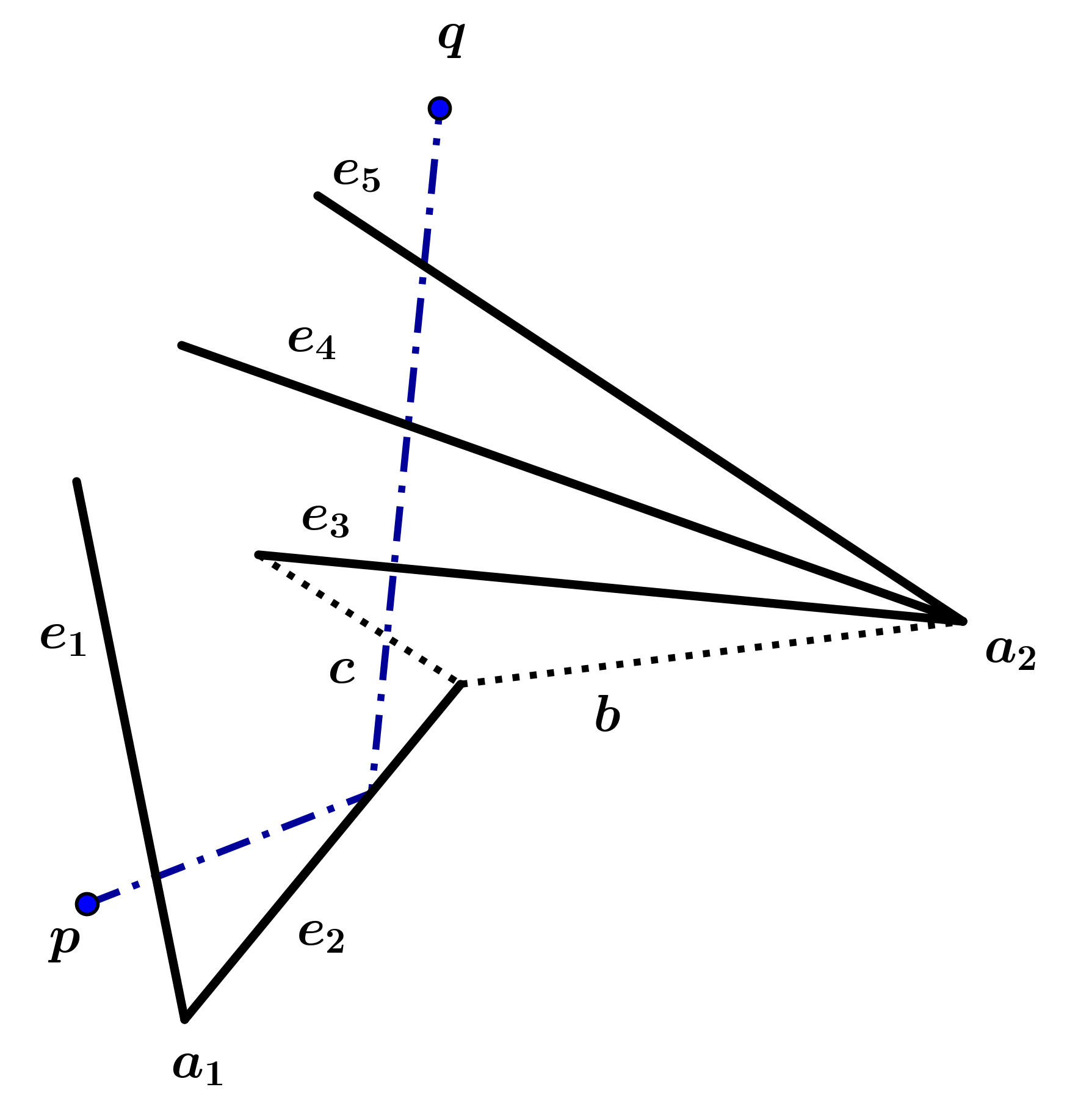}
\caption*{(i)}
\end{subfigure}
\begin{subfigure}[t]{.48\textwidth}
\centering
\includegraphics[scale=0.35]{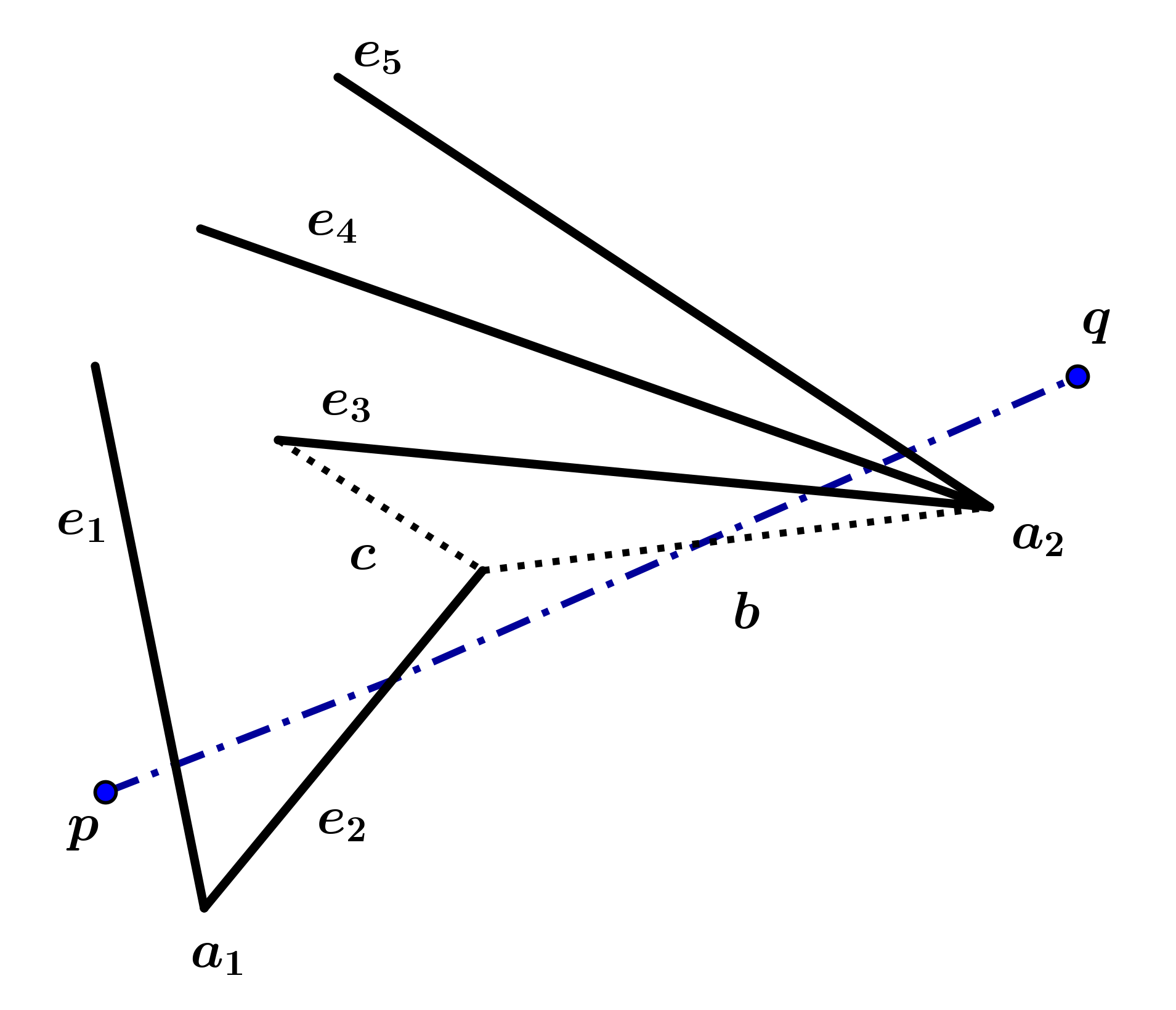}
\caption*{ (ii)}
\end{subfigure}
\caption[Funnel technique cannot be used for the SP problem for bundles]{(i): If we add $b$ into the sequence of diagonals to use the funnel technique, we cannot obtain the shortest path since the path does not pass through $b$; (ii):  if we add $c$ into the sequence of diagonals to use the funnel technique, we cannot obtain the shortest path since the path does not pass through $c$.}
\label{fig:funnel-cannot}
\end{figure}

We know that,  in the SP problem for bundles of line segments, if line segments are sorted as a list of diagonals of a sequence of adjacent triangles, the funnel technique of Lee and Preparata~\cite{Lee1984} can be applied to get the exact shortest path. However, the  domain containing the paths that we consider is not a triangulated polygon. 
In general, the shortest path along a sequence of line segments is different from the shortest path inside a simple polygon because the former can reach a line segment and reflect back.
For example, in Figure~\ref{fig:funnel-cannot}, to create a sequence of diagonals of a triangulated domain from line segments of two bundles $a_1$ and $a_2$, we have to add either the edge $b$ or $c$ but not both before using the funnel technique. But we cannot know exactly which edge should be added. 
Thus, in the case of two bundles, we have to construct two sequences of adjacent triangles and then use the funnel technique for each sequence of adjacent triangles. Since the number of combination cases is increasing with the number of bundles, the number of sub-problems to be solved is very large and equal to $2^{N-1}$, where $N$ is the number of bundles of the sequence.
Overall, the funnel technique is not efficient for solving the SP problem for bundles of line segments.

\begin{figure}[htp]
\centering
\includegraphics[scale=0.7]{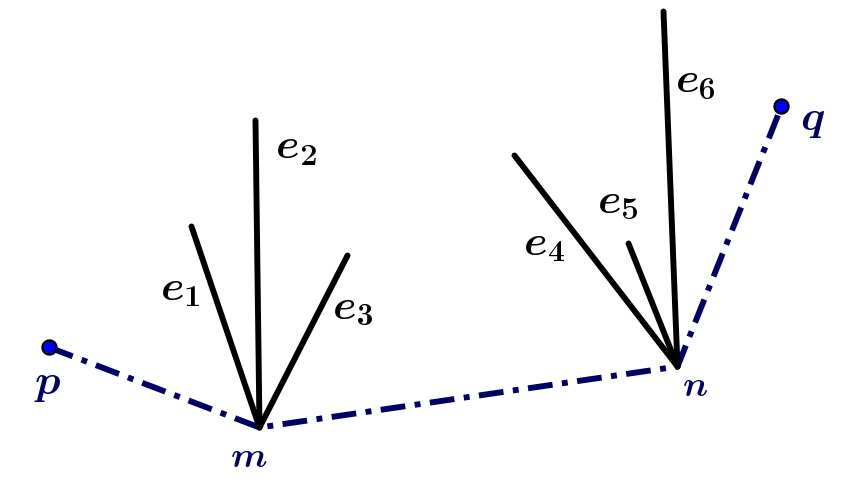}
\caption{
The technique of Li and Klette~\cite{Li2011} and the iterative method of Trang et. al.~\cite{Trang2017} cannot continue if   the obtained path at some iteration is the polyline joining $p,m,n$ and $q$, where its sub-shortest paths joining $e_1$ and $e_2$; $e_2$ and $e_3$; $e_4$ and $e_5$; $e_5$ and $e_6$  are  singletons $m$ and $n$.}
\label{fig:funnel_Li}
\end{figure}

The SP problem for pairwise disjoint line segments can be dealt with by geometric algorithms using approximate approaches by iterative steps of Li and Klette~\cite{Li2011} and Trang et. al.~\cite{Trang2017}.
%
These  algorithms  create polylines which are formed by sub-shortest paths joining two points on two consecutive line segments. After each update step, these points on line segments are adjusted  to get an approximate shortest path. 
These algorithms are not suitable to use when line segments are not pairwise disjoint
such as when the set of line segments is a sequence of bundles.
Indeed, at some iteration, if pairs of points of the obtained polyline  which are in pairs of line segments  are identical, the corresponding sub-shortest paths degenerate to  singletons. Then the adjustment for points on line segments is not possible, see Figure~\ref{fig:funnel_Li}.

\end{document}